\documentclass{article}

\usepackage{microtype}
\usepackage{graphicx}
\usepackage{subfigure}
\usepackage{booktabs} % for professional tables

\usepackage{hyperref}

\usepackage{multirow} % added by Shahriar

\usepackage{amsmath,amsfonts,bm}

\def\eqref#1{equation~\ref{#1}}
\def\1{\bm{1}}

\DeclareMathAlphabet{\mathsfit}{\encodingdefault}{\sfdefault}{m}{sl}
\SetMathAlphabet{\mathsfit}{bold}{\encodingdefault}{\sfdefault}{bx}{n}

\newcommand{\R}{\mathbb{R}}

\newcommand{\bfE}{{\bf E}}

\newcommand{\bfI}{{\bf I}}

\newcommand{\bfP}{{\bf P}}
\newcommand{\bfQ}{{\bf Q}}

\newcommand{\bfW}{{\bf W}}

\newcommand{\bfb}{{\bf b}}

\newcommand{\bfs}{{\bf s}}
\newcommand{\bfx}{{\bf x}}
\newcommand{\bfy}{{\bf y}}

\newcommand{\bfq}{{\bf q}}

\newcommand{\bfd}{{\bf d}}

\newcommand{\bfw}{{\bf w}}

\newcommand{\bft}{{\bf t}}

\newcommand{\hf}{{\frac 12}}

\newcommand{\bfepsilon}{{\boldsymbol \epsilon}}
\newcommand{\bftheta}{{\boldsymbol \theta}}

\newcommand{\bfomega}{{\boldsymbol \omega}}

\usepackage[accepted]{modified_icml2024}

\usepackage{amsmath}
\usepackage{amssymb}
\usepackage{mathtools}
\usepackage{amsthm}
\usepackage{rotating} % Added by Shahriar
\usepackage{longtable} % Added by Shahriar 
\usepackage[capitalize,noabbrev]{cleveref}

\theoremstyle{plain}

\theoremstyle{definition}

\theoremstyle{remark}

\usepackage[textsize=tiny]{todonotes}
\icmltitlerunning{Deep Optimal Experimental Design}

\begin{document}

\twocolumn[
\icmltitle{Deep Optimal Experimental Design for Parameter Estimation Problems}
% It is OKAY to include author information, even for blind
% submissions: the style file will automatically remove it for you
% unless you've provided the [accepted] option to the icml2024
% package.

% List of affiliations: The first argument should be a (short)
% identifier you will use later to specify author affiliations
% Academic affiliations should list Department, University, City, Region, Country
% Industry affiliations should list Company, City, Region, Country

% You can specify symbols, otherwise they are numbered in order.
% Ideally, you should not use this facility. Affiliations will be numbered
% in order of appearance and this is the preferred way.
\icmlsetsymbol{equal}{*}

\begin{icmlauthorlist}
\icmlauthor{Md Shahriar Rahim Siddiqui}{to}
\icmlauthor{Arman Rahmim}{to}
\icmlauthor{Eldad Haber}{goo}
\end{icmlauthorlist}

\icmlaffiliation{to}{Department of Physics and Astronomy, University of British Columbia, Vancouver, Canada}
\icmlaffiliation{goo}{Department of Earth, Ocean, and Atmospheric Sciences, University of British Columbia, Vancouver, Canada}

\icmlcorrespondingauthor{Md Shahriar Rahim Siddiqui}{shahsiddiqui@phas.ubc.ca}
\icmlcorrespondingauthor{Eldad Haber}{eldadhaber@gmail.com}

% You may provide any keywords that you
% find helpful for describing your paper; these are used to populate
% the "keywords" metadata in the PDF but will not be shown in the document
\icmlkeywords{Optimal Experimental Design, parameter estimation, deep learning}

\vskip 0.3in
]
% \printAffiliationsAndNotice{\icmlEqualContribution}
\printAffiliationsAndNotice{} % otherwise use the standard text.
\begin{abstract}
Optimal experimental design is a well studied field in applied science and engineering. Techniques for estimating such a design are commonly used within the framework of parameter estimation. Nonetheless, in recent years parameter estimation techniques are changing rapidly with the introduction of deep learning techniques to replace traditional estimation methods. This in turn requires the adaptation of optimal experimental design that is associated with these new techniques. In this paper we investigate a new experimental design methodology that uses deep learning. We show that the training of a network as a Likelihood Free Estimator can be used to significantly simplify the design process and circumvent the need for the computationally expensive bi-level optimization problem that is inherent in optimal experimental design for non-linear systems. Furthermore, deep design improves the quality of the recovery process for parameter estimation problems. As proof of concept we apply our methodology to two different systems of Ordinary Differential Equations. 
\end{abstract}

\section{Introduction}
\label{sec:sec1}

Mathematical models that are based on ordinary differential equations (ODEs) are used in various fields such as biology \cite{PrincCompBio}, ecology \cite{ecologyInvProblems}, physiology
\cite{physiologyInvProblems}, pharmacometrics \cite{pharmacometricsBook}, climate modeling \cite{keane2017climate} and more \cite{optcontrolBook, BetJ:01}. These models
often involve unknown parameters that need to be estimated
from experimental data. The parameters represent rates and constants that are usually important in understanding the dynamics of the underlying system, and could be crucial to decision making and to system's control. Examples range from determining the insulin resistance of an individual from the clinically used intravenous glucose tolerance test (IVGTT) \cite{ChungHaber12}, finding the
rates of transfer of radio-pharmaceuticals and 
determining when a system will exhibit chaotic behavior \cite{hun07a}.   

%A significant example we consider here is from the field
%of nuclear medicine, called the ”3 Tissue-Compartment” 
%(3-TC) model \cite{bitker2022non}, which is an Ordinary Differential Equation
%(ODE) system. 
Parameter identification is typically performed in two steps: data 
collection and data fitting. In the data collection stage
experiments are performed and data are measured. In the data
fitting stage, a data fitting procedure is used to evaluate
parameters from the (typically noisy) measurements. For this process to be effective, the experiment must
be carefully designed. 

In many cases one can densely measure the data
yielding a grossly over-determined problem \cite{Bjo96a}, however, in other cases, the measurement process is difficult and expensive, and this leads to restrictions on the type,  amount and quality (that is, signal to noise) of the data that is collected.
A decision must be made
about which data should be measured, at what frequency and to what accuracy.
A consequent trade-off exists
between the accuracy of parameter recovery and the amount and quality of the data. 
The
goal of this paper is to propose an experimental design technique that
balances parameter estimation and the
cost of the experiment.  

{\bf Previous Work:}
Optimal experimental design is an important topic in applied science and engineering with applications in medical imaging, geoscience, optimal control, public health and many other fields (see \cite{bardow2008, Pukelsheim93, allaubjou, ChalonerVerdinelli1995, HaberHoreshTenorio08,AtkinsonDonev1992} and references therein). 

In this paper we focus our attention to the design of experiments for nonlinear systems that are governed by ODEs, with the aim to identify parameters 
within the equations. While there is significant work on the topic (see \cite{HaberHoreshTenorio09, ChungHaber12, mclellan, banga2008parameter,bock2013parameter} and references therein)
the methods proposed for the solution of the problem are difficult to apply and require classical recovery techniques.

\begin{figure}
    \centering
    \includegraphics[width=7cm]{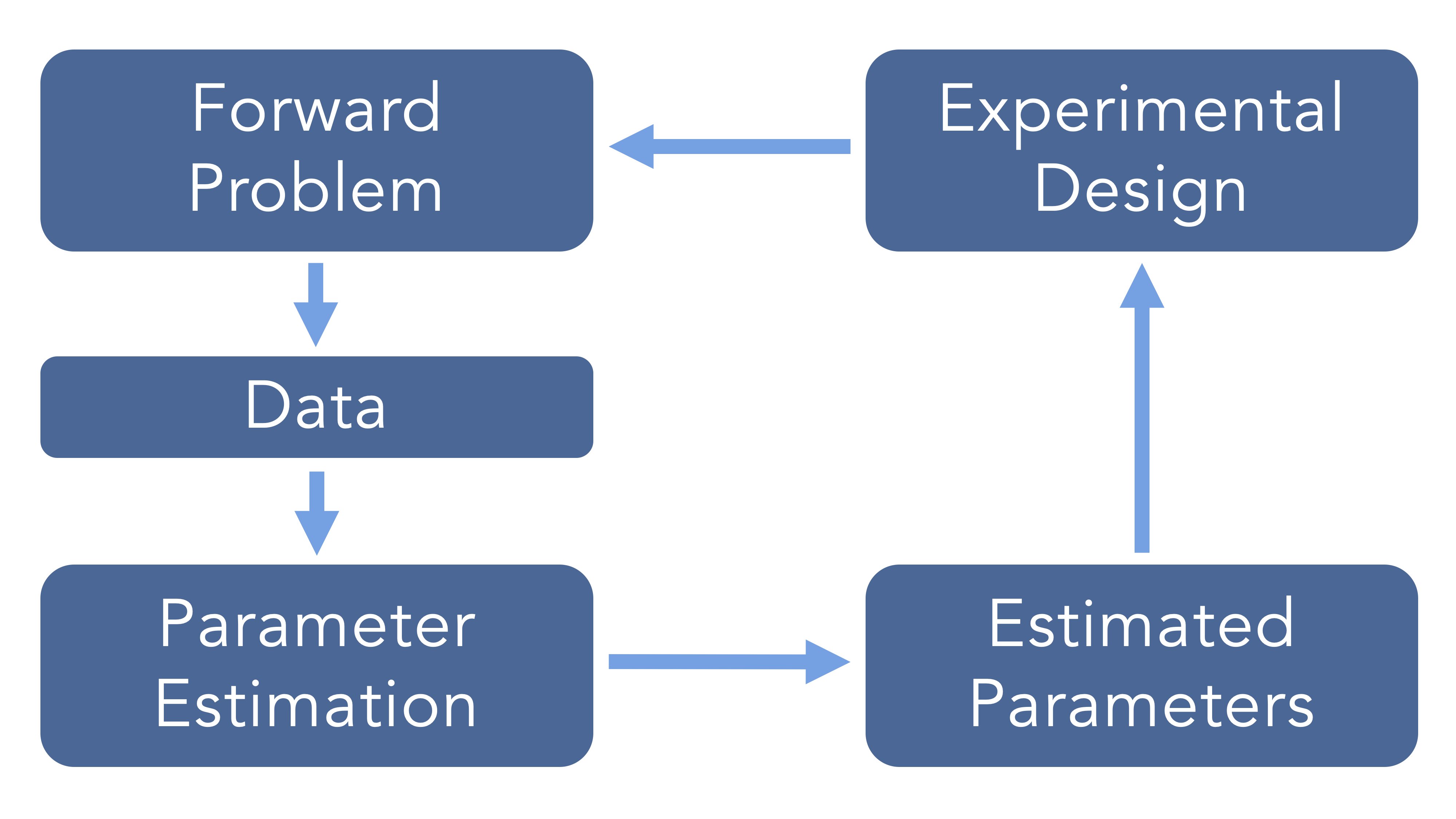}
    \caption{A schematic process of parameter estimation and experimental design. Given a set of plausible parameters and a particular experimental setting the data of the forward process is simulated. One then use some parameter estimation routine and measures the quality of the estimated parameters.  The design is changed to have better estimation of the parameters.}
    \label{fig:fig1}
\end{figure}

The design process is based on three pillars. 
In the core stands the ability to quickly solve the so called forward problem for different experimental settings and parameters. In our case, the forward problem is a discretized system of ordinary  differential equations that needs to be solved many (in some cases hundreds of thousands) of times.  The second pillar of the design is the solution of the so-called inverse problem. That is, the identification of parameters given some experimental settings. For linear problems with quadratic regularization (that is, Gaussian priors), it is possible to obtain an analytic expression for this problem (see \cite{HaberHoreshTenorio08}). However, for nonlinear inverse problems such expressions are impossible to obtain and therefore, numerical optimization techniques are typically used for the solution of the problem and estimating the parameters. The solution of the inverse problem is sometimes referred to as the inner optimization problem.  Finally, in the third and outer pillar of the design process stands an optimization problem that aims to obtain the best parameter recovery (on average) from the inner optimization problem by changing the experimental setting. The process is illustrated in Figure~\ref{fig:fig1}.
\vspace{-5 pt}

Optimal experimental design for nonlinear systems is therefore a bilevel optimization problem.
The process of solving the outer problem (the design problem) is iterative. For a given design, one chooses some parameters to be identified from some distribution and solves many inner optimization problems to estimate those parameters. Comparing the parameters to their true values and taking derivatives it is possible to obtain a direction that improves the recovery of those parameters.
Design methods for parameter estimation differ in 
their outer objective function and the statistical framework, Bayesian vs Frequentist \cite{ChalonerVerdinelli1995, HaberHoreshTenorio08}. Nonetheless, all methods known to us rely on repeated solution of the inverse problem, for different parameters, which makes the problem very difficult to solve and computationally challenging.
To this end, a few authors have proposed to replace the forward solution with reduced models \cite{wang2023optimal, ushijima2015experimental}; however those can behave poorly, especially when changing the design.

In recent years deep learning has been used in the context of generative models that are able to sample the main modes of a distribution (see \cite{yang2023diffusion}
and references therein). The ability to learn a distribution or its main mode can be very useful when considering optimal design. 
In particular, we consider Likelihood Free Estimators (see e.g. \cite{sainsbury2022fast, didelot2011likelihood, papamakarios2019neural_1, papamakarios2019neural_2, papamakarios2019neural_3}) that allow for the estimation of the solution of the inverse problem without the solution of the forward problem. With improvements in neural network architectures,  such estimators have become very attractive for problems where the solution of the forward problem is computationally challenging. 

{\bf The novelty of this work } is the development of a methodology that uses Likelihood Free Estimators for optimal experimental design for a system governed by differential equations. Our method  circumvents the  bilevel optimization
problem that is traditionally used. 
We propose a mathematical framework that allows for the construction of a  deep network that yields an optimal recovery of the parameters and the experimental setting. Given a parameter to be recovered and a forward problem, our framework requires the ability to sample from the prior distribution of the parameter and to solve the forward problem for those samples. Our framework  does not require the solution of the inverse problem via optimization and rather learns a direct map from the data to
the parameters under different experimental settings. By using an appropriate loss function, we are able to find a network that is optimal for the recovery of the parameter and the optimal experimental settings.

The rest of this paper is organized as follows. In Section~\ref{sec:sec2} we give a mathematical background to the field of design in the context of differential equations. In \Cref{sec:sec3}
we discuss Likelihood Free Estimators and show how they can be used for the design problem. In Section~\ref{sec:sec4} we propose two training methods to train such estimators. In Section~\ref{sec:sec5} we conduct a number of experiments with systems from diverse fields that show that our method is robust. Finally, in Section \ref{sec:sec6} we summarize the paper.

\section{Mathematical Background for Optimal Design}
\label{sec:sec2}

Consider a parameter estimation problem, and let $\bfq \in {\cal Q}$ be a $p$ dimensional parameter vector in a differential equation. Let $\bfomega \in \Omega$ be a set of $e$ parameters that represent some experimental settings. Finally,  let $\bfd \in {\cal D}$ be the data measured. Formally, we write
\vspace{-3 pt}
\begin{eqnarray}
    \label{forward}
    \bfd = F(\bfq, \bfomega) + \bfepsilon
\end{eqnarray}
Here $F: {\cal Q}\times \Omega \rightarrow {\cal D}$ is the forward problem that takes the parameter $\bfq$ and experimental setting $\bfomega$ and yields some measured data, $\bfd$.
In our context, the application of $F$ requires the solution of a system of ODEs, typically an initial value problem.
The data are assumed to be polluted with random noise $\bfepsilon \sim \mathcal{N}(0, \sigma^2\bfI)$ which is assumed to be normal with $0$ mean and $\sigma^2$ standard deviation.
We assume that $\bfq$ is associated with a known probability density function $\bfq \sim \pi(\bfq)$. In practice one does not require to have a mathematical expression for the  density, however, we assume to have sufficient amount of samples, $\bfq_i, i=1, \ldots, N_q$, that can be generated from this density.
The experimental setting vector $\bfomega$ is assumed to be under our control. Such a vector may represent measurement times, frequencies for the forward problem, source terms or other controllable parameters in the experiment.

A common approach to estimate the parameter $\bfq$ given the data $\bfd$ is to use Bayesian inference. Using Bayes' theorem, the probability density function of $\bfq$ given $\bfd$ is 
\begin{eqnarray}
    \label{bayes1}
  \pi_{\bfomega}(\bfq | \bfd) &\propto& \pi(\bfq) \pi_{\bfomega}(\bfd|\bfq) \\
  \nonumber
  &=& 
  \pi(\bfq) \exp\left( -{\frac {1}{2\sigma^2}} \| \bfd - F(\bfq; \bfomega) \|^2 \right).
\end{eqnarray}
where $\pi(\bfd|\bfq; \bfomega)$ is the likelihood, that is, the probability of the data, $\bfd$ given the parameter $\bfq$ and the experimental setting $\bfomega$. 
The Maximum A Posteriori (MAP) estimate is the parameter that maximizes this distribution; that is
\begin{eqnarray}
    \label{map}
    \widehat \bfq(\bfomega) = {\rm arg}\min_{\bfq} {\frac 1{2\sigma^2}} \|F(\bfq, \bfomega) -\bfd \|^2 + R(\bfq) 
\end{eqnarray}
(assuming there is a unique global minimizer).
Here $R(\bfq) = -\log(\pi(\bfq))$ is the negative log of the prior.

Associated with any estimator $\widehat \bfq$ is the recovery loss or the {\bf risk} defined by
\begin{eqnarray}
    \label{rloss}
    \ell_q(\bfomega) = 
    {\mathbb E}_{\bfq, \bfepsilon}\, \hf\|\widehat \bfq(\bfomega) - \bfq \|^2
\end{eqnarray}
Note that the risk is obtained by taking the expectation over the parameter $\bfq$ and the noise $\bfepsilon$, and therefore, the risk depends on the experimental setting alone. 
Associated with the estimator $\widehat \bfq$ is the estimated data 
\begin{eqnarray}
\label{estdata}
\widehat \bfd = F(\widehat \bfq, \bfomega)    
\end{eqnarray}
We also consider the data risk
\begin{eqnarray}
    \label{datarisk}
    \ell_d(\bfomega) = {\frac 1{2n}} {\mathbb E}_{\bfq, \bfepsilon}\, \sum \|F(\widehat \bfq, \bfomega_i) - \bfd_i \|^2.
\end{eqnarray}
When $n$ are the total number of potential measurements and which allows the application of non-uniform time intervals. 

Note that two possible estimated solutions $\widehat \bfq$ with similar risk $\ell_{q}(\bfomega)$ may have different data risk
$\ell_d(\bfomega)$. We therefore consider the total loss as a weighted sum of the risk \eqref{rloss}
and the data risk \eqref{datarisk}, 
that is
\begin{eqnarray}
    \label{totalrisk}
    \ell_{T}(\bfomega) = \ell_q(\bfomega) + \gamma \ell_d(\bfomega).
\end{eqnarray}
where $\gamma$ is a hyper-parameter (we chose $\gamma=1$ in our experiments). 
Using the definitions \eqref{totalrisk} allows for a framework for the optimal design. In particular let 
\begin{eqnarray}
    \label{oed}
    \bfomega^* \in {\rm arg}{\min}_{\bfomega}\, \ell_T(\bfomega)
\end{eqnarray}
The solution $\bfomega^*$ is interpreted as the experimental setting that yields the best recovery of the parameter, $\bfq$, on average that also fit the measure data.

One main difficulty in solving \eqref{oed} is that the problem may not be differentiable with respect to the experimental setting $\bfomega$.
A common way to overcome this difficulty that has been reviewed in \cite{boydBook}  is to discretize the space of experimental setting, $\Omega$, finely obtaining many plausible experiments $[F(\bfq, \bfomega_1), \ldots, F(\bfq, \bfomega_s)]$, that can be done. 
Assume that we have $s$ possible different experiments and assume for simplicity that each datum $\bfd_i$ corresponds to the data that is recorded from a different experiment $\bfomega_i$.
Introducing weights $0 \le \bfw$, one replaces the original posterior \eqref{bayes1} with one that contains all the data for all possible experiments, weighted by $\bfw$, that is
\begin{eqnarray}
    \label{bayes2}
   \pi_{\bfw}(\bfq | \bfd) \propto
  \pi(\bfq) \exp\left( {-\frac {1}{2\sigma^2}} \| \bfw \odot ( \bfd - F(\bfq) ) \|^2 \right).
\end{eqnarray}
where $F(\bfq)$ assumes that all possible experiments are conducted.
The problem of estimating $\bfomega$ is replaced with the estimation of the weights $\bfw$. Clearly, if $\bfw$ is sparse then only a few experiments are to be conducted. Therefore, it has been proposed
\cite{HaberHoreshTenorio08} to replace the original problem with a penalized problem
\begin{eqnarray}
    \label{oedw}
    \bfw^* = {\rm arg}{\min}_{\bfw}\, \ell_T(\bfw) + \alpha\,  {\rm Sp}(\bfw)
\end{eqnarray}
where ${\rm Sp}(\bfw)$ promotes sparsity in $\bfw$, and $\alpha$ is a hyper-parameter. A common approach is using the 1-norm although using approximations to zero-norm have been proposed \cite{donoho04sparsest,eladReview}.

At this point it is worth while exploring the "meaning" of the weights, $\bfw$. It is interesting to note that different statistical framework implies a different meaning for the weights. A standard approach  presented in \cite{boydBook} views the term $\pi_{\bfw}$ in \eqref{bayes1} as the likelihood and therefore $\bfw_i$ is interpreted as the inverse standard deviation of the $i$-th datum. This implies that $\bfw_i$ represents the accuracy in which we are supposed to measure the data. If $\bfw_i \approx 0$ then the data can be recorded with infinite errors, which implies that it is not needed. A second interpretation, discussed in \cite{HaberHoreshTenorio08} is that $\bfw$ are just weights used algebraically within the recovery process and do not have real statistical meaning. Finally, another common approach is to assume that we do not have control over the accuracy of the measured data and that each datum has a given standard deviation. In this case $\bfw$ in a binary variable. Regardless of the statistical framework taken, the mathematical problem to be solved is similar and in the rest of the paper we discuss efficient methods for its solution.

Even after the reformulation of the problem,
solving the optimization \eqref{oedw}  is very difficult. It requires solving the optimization problem \eqref{map} for many different $\bfq$'s
in order to estimate the risk $\ell_T(\bfw)$. This in turn requires solving many forward problems which can be very expensive especially in the context of ODEs.
Furthermore, computing derivatives with respect to $\bfw$ is challenging and requires implicit differentiation.

Therefore, in the next section we describe an alternative to this process that leads to a much more efficient algorithm.

\section{Likelihood Free Estimators and Optimal Design }
\label{sec:sec3}

The optimization problem \eqref{oedw} is difficult because we require to estimate $\ell_T(\bfw)$ which in turn requires the solution of the optimization problem \eqref{map}. The optimization stems from our use of the MAP estimator. 
The MAP estimator can be thought of as a nonlinear function of the form
\begin{eqnarray}
    \label{mapest}
    \widehat \bfq = F^{\dag}_{\rm map}(\bfw \odot \bfd)
\end{eqnarray}
where the nonlinear function $F^{\dag}_{\rm map}$ maps the data to an estimator $\widehat \bfq$, where the Hadamard product $\odot$ represents element-wise multiplication.

The MAP estimator can be highly useful and has very desirable properties. However, it may be far from optimal. If the posterior is highly skewed, then the MAP may not lead to the minimization of the risk \eqref{rloss}. In practice, the main reason that the MAP estimator is commonly used is  our ability to compute it. 
In many cases, estimators such as the conditional mean may be attractive; however they require non-trivial computations and Monte-Carlo integration. 
Thus, the MAP estimator is certainly not the only estimator that can be used. In the context of learning it is possible to directly learn an estimator that minimizes the risk~\eqref{rloss}. In many cases, such an estimator can perform even better than the MAP estimator (see for example \cite{papamakarios2019neural_1, papamakarios2019neural_2, papamakarios2019neural_3}). Furthermore, by appropriately setting its architecture, it is possible to include different experimental setting and noise levels in this estimator and then use it for the solution of the optimal design.

%\subsection{Likelihood Free Estimators}

We now derive a Likelihood Free Estimator for the solution of the problem, that enable us to solve both the estimation problem as well as the design problem.
To this end, let us define the estimator
\begin{eqnarray}
    \label{apinv}
    \widehat \bfq((\bfomega) = F^{\dag}_{\bftheta}( \bfw \odot \bfd, \bfw, \sigma). 
\end{eqnarray}
The estimator depends on 
the experimental design vector $\bfw$ that controls the design and trainable parameters $\bftheta$. It also uses the parameter $\sigma$ that represents the noise level.
Note that this estimator does not require solving the forward problem directly and computing the likelihood, hence its name, Likelihood Free Estimator (LFE).

The risk for this estimator can be written as
\begin{eqnarray}
    \label{rloss_theta}
    \ell_T(\bftheta, \bfw) = 
    {\mathbb E}_{\bfq, \bfepsilon} \|F^{\dag}_{\bftheta}(\bfw \odot \bfd, \bfw, \sigma) - \bfq \|^2 + \gamma \ell_d(\bftheta, \bfw)
\end{eqnarray}
Clearly, the best estimator is obtained by minimizing the risk with respect to weights $\bftheta$ and the experimental setting $\bfw$. 
It is important to note that since the estimator is a simple function evaluation, we can solve a single optimization problem to estimate both the estimator parameter $\bftheta$ and the experimental design parameters $\bfw$.  Thus, this formulation avoids the need for bilevel optimization all-together.

%\subsection{Network Architecture}

We now discuss a proposed architecture that we use
for the solution of the problem. The architecture
uses the data, $\bfd$, the experimental setting $\bfw$
and the noise level $\sigma$ in order to compute the estimator $\widehat \bfq$. A minimal structure of the network is summarized in
Algorithm~\ref{alg:net}, where the data vector $\bfd$ is assumed to be of length $N$. 
% \begin{algorithm}[h]
%    \caption{Neural Network Architecture for Design}
%    \label{alg:net}
% \begin{algorithmic}
%     \INPUT $\bfd, \sigma$
%    \STATE Set $\bfy = \bfQ_0(\bfw \odot \bfd)$
%    \STATE Embed $\bfs = \bfE(\sigma), \bfz = \bfW_0\bfw$
%    \STATE Generate the augmented tensor 
%    $\bfx_0 = [\bfy, \bfz, \bfs]$
%    \FOR{$i = 0,...,{\rm (nlayers-1)}$}
%    \STATE  $\bfx_{i+1} = \bfx_i + \bfK_i \eta(\bfQ_i \bfx_i + \bfb_i)$
%    \ENDFOR
%     \STATE $\widehat \bfq = \bfQ_f \bfx $
%    \STATE {\bf return} $\widehat \bfq$
% \end{algorithmic}
% \end{algorithm}
\begin{algorithm}[h]
   \caption{Neural Network Architecture for Design}
   \label{alg:net}
\begin{algorithmic}
    \INPUT $\bfd, \sigma, \text{n\_layers}$
    \STATE Initialize weight vector $\bfw = \text{torch.ones}(N)$
    \STATE Set $\bfw = \text{sparsify}(\bfw)$
    \STATE Compute initial data embedding $\bfy = \bfW_1(\bfw \odot \bfd)$
    \STATE Embed $\bfs = \bfE(\sigma), \bfq = \bfQ_1(\bfw)$
    \STATE Apply nonlinearity $\bfq = \eta(\bfq), \bfs = \eta(\bfP_1(\bfs))$
    \STATE Generate the augmented tensor 
    $\bfx_0 = \eta(\bfy + \bfq + \bfs)$
    
    \FOR{$i = 1,..., {\rm n\_layers}$}
       \STATE Compute new layer $\bfx_{i} = \bfW_{i+1}\bfx_{i-1}$
       \STATE Update $\bfq = \eta(\bfQ_{i+1}(\bfq)), \bfs = \eta(\bfP_{i+1}(\bfs))$
       \STATE Update the tensor $\bfx_{i} = \eta(\bfx_{i} + \bfq + \bfs)$
    \ENDFOR
    
    \STATE Compute final output $\widehat \bfq = \bfW_f(\bfx_{n\_layers})$
    \STATE {\bf return} $\widehat \bfq$
\end{algorithmic}
\end{algorithm}

% \textcolor{blue}{The network is a modified feedforward neural network with residual connections, consisting of weights $\bftheta = {\bfW_i, \bfQ_i, \bfP_i}$, where $i = 1, \ldots , n\_layers}$. A few key differences distinguish this network from standard residual architectures. The first difference is that the input data $\bfd$ is weighted with the parameters $\bfw$,
% allowing us to control the experimental design and train for the optimal parameters. Second, the parameters $\bfw$ are embedded into the network. This allows for the network to be aware of the experimental setting. Finally, similar to diffusion models \cite{li2018diffusion}, we embed the noise into the network, allowing it to be trained for different noise levels. In the network $\eta$ stands for the activation function 
% bundled with possible layer normalization. }
The network is a simple residual network 
with weights
$\bftheta = \{\bfW_0,\bfQ_0, \bfQ_f, \bfQ_i, \bfP_i, \bfb_i\}, i=0,..., {\rm (nlayers-1)}$ 
with a few important differences from standard residual networks. The first difference is that the input data $\bfd$ is weighted with the parameters $\bfw$,
allowing us to control the experimental design and train for the optimal parameters. Second, the parameters $\bfw$ are embedded into the network. This allows for the network to be aware of the experimental setting. Finally, similar to diffusion models \cite{li2018diffusion}, we embed the noise into the network, allowing it to be trained for different noise levels.
In the network $\eta$ stands for the activation function 
bundled with possible layer normalization. 

The network above can be modified to have different embedding for $\sigma$ and $\bfw$  at each layer. However,
we have found that even this simple architecture yields very good results.
The network needs to be trained for its weights as well as for the best experimental settings, $\bfw$. This is discussed in the next section.

\section{Training the Estimator}
\label{sec:sec4}

In this section we discuss training the network by using
a self-supervised approach \cite{jaiswal2020survey,zhai2019s4l}. Such techniques have been the cornerstone of recent advances in the application of deep learning for realistic application.
However, before we proceed, we need to discuss the particular parametrization used for this problem.
While choosing the network weights is standard, there
are two options for the choice of the design vector $\bfw$. In the first option, we allow $\bfw$ to have any real non-negative value. This is the simplest case 
as it allows us to simply train the experimental setting together with the parameters of the network, estimating
the network and the optimal design simultaneously.
In the second option we treat $\bfw$ as a binary variable. Binary variables are considered where we either conduct the experiment or not implying that the data is measured or not. In this case, training the network implies solving a mixed nonlinear-binary programming problem. We solve this problem by a combination of Tabu-search for the binary variables and
stochastic gradient descent for the continuous variables.

Regardless to the optimization process, we use a self supervised approach. That is, the algorithm generates its own data and trains on it. A summary of the algorithm is given in Algorithm~\ref{alg:sst}.
\begin{algorithm}[h]
   \caption{Training an LFE estimator}
   \label{alg:sst}
\begin{algorithmic}
    \INPUT Network, Sampling routine for $\bfq$
   \FOR{$i = 0,...$}
   \STATE  Sample a batch $\bfq$
   \STATE Use the forward problem to compute $\bfd = F(\bfq) + \bfepsilon$
   \STATE Use the network to recover $\widehat \bfq$
   \STATE Compute $\widehat \bfd = F(\widehat \bfq)$
   \STATE Compute the loss in \eqref{oedw}
   \STATE Update parameters to minimize the loss
   \ENDFOR
\end{algorithmic}
\end{algorithm}

The algorithm is self trained in a sense that it continuously samples $\bfq$ and computes its corresponding data, and then uses the data in order to estimate $\bfq$. The algorithm requires a way to sample from the prior distribution of $\bfq$. For many parameter estimation problems such a distribution is readily available. In the examples below we sample the parameters from a log-normal distribution.
The ability to generate an infinite amount of data, allows for a very robust training procedure in which over-fitting does not play a role.

The algorithm~\ref{alg:sst} minimizes the loss with respect to the design variables $\bfw$ and the network parameters $\bftheta$. However, updating the design variables changes, depending on their type.
We now discuss the two approaches used for the update of the design variables.

\subsection{Continuous Design Variables (Method 1)}
\label{cont_design_section}
Consider first the case where the design vector $\bfw$ can take on any values $\R_{>0}$. 
In this case we seek to minimize the recovery loss $\ell_T(\bfw)$ and enhance the sparsity of $\bfw$.

A simple algorithm to achieve that is to use the 
soft shrink function \cite{eladReview}.
Here we modify the classical soft shrink function to keep non-negativity.
The modified soft shrink function is defined as
\begin{eqnarray}
    \label{sh}
    s_{\rho}(t, \rho) = \left\{\begin{matrix} t -\rho & t>\rho \\ 0   & {\rm otherwise}
    \end{matrix}
    \right.
\end{eqnarray}
\vspace{-1 pt}This function is used during minimization in a straight forward manner (see \cite{eladReview} for convergence proofs). Using the stochastic gradient descent 
algorithm we update $\bftheta$ in a standard way but for $\bfw$ we use the following formula
\vspace{-8 pt}  % Adjust the length as needed
\begin{eqnarray}
    \label{sh}
    \bfw \leftarrow s_{\rho}(\bfw - \mu \delta \bfw, \rho)
\end{eqnarray}
Here the soft shrink function is applied element-wise, 
$\mu$ is the learning rate and $\delta \bfw$ is the descent direction, obtained by the stochastic gradient descent algorithm.
\subsection{Binary $\bfw$ and Tabu Search (Method 2)}
\label{binary_design_section}
Using continuous weights $\bfw$ is attractive as it enables the use of continuous optimization techniques. 
Nonetheless, the result of such an algorithm can lead to
many weights that are small, raising the question about the importance of the data that is associated with those weights.
In many applications, a binary $\bfw$ makes more sense.
In this case, $\bfw_i=1$ implies that the data is measured and when $\bfw_i=0$ the data is ignored and not measured. Thus, minimizing the loss $\ell_T$ is involved with mixed nonlinear binary programming, which is a challenging problem.
To approximate the solution of this problem we combine a Tabu search method (for the binary variables) and a standard stochastic gradient descent for the continuous variables. Tabu search methods are commonly used in binary programming \cite{hertz1995tutorial}. The algorithm uses the current iterate to define neighbors.
These neighbors are tested, and the algorithm steps into the lowest one. The algorithm retains a list (the Tabu list) of points that are visited in order to not repeat points. Although the algorithm is very simple, it finds reasonable approximations to the minimum of binary programming problems (see \cite{gendreau2005tabu} for details). To use Tabu within the context of the optimization problem we incorporate it in a block coordinate descent algorithm. The algorithm is summarized in 
Algorithm \ref{alg:sst_binary}.
\vspace{-6 pt}  % Adjust the length as needed
\begin{algorithm}[h]
   \caption{Training LFE with binary variables}
   \label{alg:sst_binary}
\begin{algorithmic}
    \INPUT Network
    \STATE Initialize $\bfw$ and $\bftheta$
   \FOR{$i = 0,..., {\rm outer\_iter}$}
   \FOR{$j = 0,..., {\rm inner\_iter}$}
   \STATE Compute the loss in \eqref{oedw}
   \STATE Compute gradients with respect to $\bftheta$
   \STATE Update $\bftheta$
   \ENDFOR
   \STATE Set the Tabu list to empty.
   \FOR{$j = 0,..., {\rm inner\_iter}$}
   \STATE Compute the neighbors for $\bfw$ 
   \STATE Compute the loss in \eqref{oedw} for all neighbors
   \STATE Update $\bfw$ to the lowest loss
   \STATE Update the Tabu list
   \ENDFOR
   \ENDFOR
\end{algorithmic}
\end{algorithm}
\vspace{-5 pt}  % Adjust the length as needed
Solving a mixed binary nonlinear programming problem is substantially more expensive compared with solving the problem with continuous variables. However, in our application we have found that a very small number of steps on the outer iteration usually suffice to obtain reasonable accuracy. 
\section{Numerical Experiments}
\label{sec:sec5}
In this section we study two different problems. The first problem is of practical importance to the field of nuclear medicine, utilizing a "3-Tissue Compartment" (3-TC) model, an ODE system amongst a range of compartmental models  in the field of kinetic modeling \cite{armanRef1_morris2004kinetic} \cite{armanRef2_bentourkia2007tracer}, to estimate physiological parameters of importance using nuclear medicine imaging. The second is an example of a non-linear system of differential equations that model the populations of predators and prey in a system known as the Lotka-Volterra predator-prey model (PPM).
We now shortly describe these problems, a detailed implementation can be found in Appendix \ref{appendix_A}.

\subsection{3-Tissue Compartment Model}
\label{3TC_section}
A multi-compartment model is a simplified (or reduced) mathematical model used to describe the movement of mass or energy between different compartments within a system. 
Such models are often used to generate a reduced yet useful model of a complex system that is difficult to represent accurately. Each compartment in the model is considered to be homogeneous, thus allowing to model only the interaction between compartments. 
Compartment models have various applications in several fields, including pharmacokinetics, epidemiology, systems theory, complexity theory, engineering, physics, and social science \cite{armanRef1_morris2004kinetic} \cite{CompModelRef2} \cite{CompModelRef3}.
Here we consider the 3-Tissue Compartment Model (3-TC) model \cite{ZakariaeiP1449} from the field of nuclear medicine and molecular imaging. The model is a system of ODEs, and is an example of a compartmental model from kinetic modeling. 
The  3-TC model (Figure \ref{fig:3comp} in the Appendix) describes the kinetics of a pharmaceutical in tissue, and how its concentration changes with time in each of the compartments. The model is written as
\begin{subequations}
\label{3-TC-equations}
\vspace{-0.58\baselineskip} % Adjust the value as needed
\begin{align}
\frac{dP_{int}}{dt}& &=& P_{v} k_{1} - (k_{2}+{k_{3}})P_{int} + P_{b}k_{4} \\
\frac{dP_{b}}{dt}& &=& P_{int}k_{3} - (k_{4}+k_{5})P_{b} \\
\frac{dP_{intern}}{dt}& &=& P_{b} k_{5} - k_{6}P_{intern}
\end{align}
\end{subequations}
The equations describe the dynamics of the three states $P_{int}, P_b$ and $P_{intern}$ which represent, respectively, the concentration of the radiopharmaceutical in the interstitial space between cells, bound to cell receptors, and finally internalized into the cells. $P_{v}$ represents the blood input function (source) describing the delivery of the radiopharmaceutical to the tissue by the vascular system, and is the instantaneous concentration of the radiopharmaceutical in the blood. This input function exhibits a sharp spike in concentration at earlier times. The dynamics of the model depends on the parameter vector $\bfq=[k_1, \ldots, k_6]$.
The parameters typically exhibit a log-normal prior distribution, that is
\begin{equation}
\label{eq:3TC_logprior}
\ln(\bfq) \sim \mathcal{N}(\mu_i, \sigma_i^2) \quad \text{for } {i = 1, 2, 3, 4, 5, 6}
\end{equation}
\begin{figure}[h]
    \centering
    \includegraphics[width=8cm]{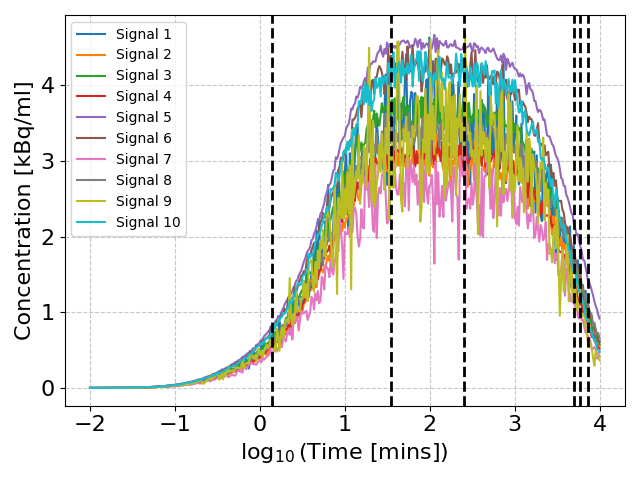}
    \vspace{-10 pt}  % Adjust the length as needed
    \caption{Synthetic Time Activity Curves for the 3-TC model at various (multiplicative) Gaussian noise levels. The vertical dotted lines correspond to an optimal data sampling scheme for $sparsity=6$ that minimizes $\ell_{T}(\bfw)$, obtained using binary design variables $\bfw$ (Method 2). $400$ time points with logarithmic spacing are considered.The design weight vector $\bfw$ has the value $1$ at the optimal time points and $0$ for the others.} 
    \label{fig:ExamplePetData_withOptPoints}
\end{figure}The parameters $\mu_{i}, \sigma_{i}$ were calculated from \cite{KlettingPaper}, and the numbers used are given in Appendix \ref{appendix_A}. Data collection is done as follows: a patient is injected with a radiopharmaceutical and is subsequently imaged with a PET/CT scanner. The imaging data comprises so-called Time-Activity-Curves (TACs), obtained from dynamic nuclear medicine imaging, that show the evolution of radioactivity concentration (kBq/ml) in the patient over time. We model these TAC signals as the sum of contributions from the 3 tissues, namely $P_{int}+P_{b}+P_{intern}$, with Gaussian noise levels $\sigma = 0\%, 1\%, 2\%...19\%$ (picked at random at each sampling). Imaging at numerous time points is costly, and an optimal design aims to identify the best times for data collection. An example of noisy TACs that were used in our experiments with the assumed multiplicative Gaussian noise at varying levels $\sigma$ is presented in Figure~\ref{fig:ExamplePetData_withOptPoints}.
\begin{figure}[h]
    \centering
    \includegraphics[width=8cm]{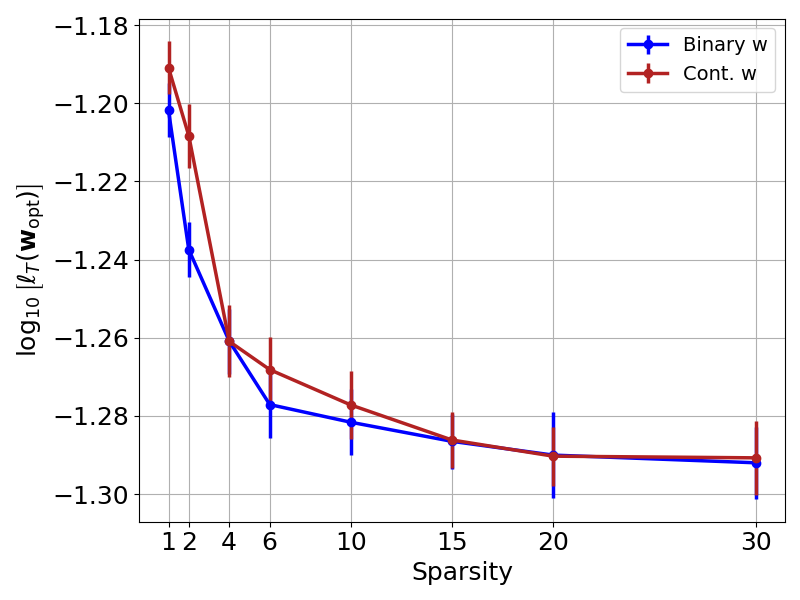}
    \caption{A comparison of the performance of the estimator trained using continuous $\bfw$ (Method 1) and binary $\bfw$ (Method 2) for the 3-Tissue Compartment (3-TC) model. Networks trained for optimal designs $\bfw_{\rm opt}$ at the sparsities shown were evaluated for $\ell_{T}(\bfw)$ over $175K$ unseen samples of $\bfq$ (and corresponding $\bfd$). Error bars showcase the Standard Error of the Mean (SEM) for the mean risk $\ell_{T}(\bfw)$ exhibited by each network at its corresponding sparsity. See Appendix \ref{appendix_B} for details.}
    \label{fig:L_curve_3TC}
\end{figure}

\subsection{Lotka-Volterra Predator-Prey Model}
\label{PPM_section}
In our second set of experiments we consider the Predator-Prey Model (PPM) which is a system of non-linear differential equations that exhibits very different behaviour from the 3-TC model. While the 3-TC model is controlled by decay, the  PPM model is cyclical in nature.
The model describes the dynamics of two species $x$ and $y$ and can be written as
\begin{equation}
\label{LotkaVolterra}
\vspace{-0.2\baselineskip} % Adjust the value as needed
% \begin{align}
\frac{dx}{dt} = \alpha x - \beta xy \quad \quad
\frac{dy}{dt} = \delta xy - \gamma y
% \end{align}
\end{equation}
equipped with some initial conditions.
The model depends on the parameter vector $\bfq = [\alpha, \beta, \gamma, \delta]$.
 The parameter $\alpha$ denotes the maximum per capita growth rate of prey, while $\beta$ signifies the rate at which prey are consumed by predators, reflecting the predator's impact on prey population reduction. Parameters $\delta$ and $\gamma$ for predators represent their natural growth rate through prey consumption and per capita death rate, respectively. The model assumes prey have an unlimited food supply, reproducing exponentially unless preyed upon. Additionally, it assumes prey are the sole food source for predators, with all environmental variables held constant.
% Changing the parameters can lead to a very large change in the dynamics. As example, we plot the phase space diagram (that is $y(t)$ vs $x(t)$) for 5 different values
% of $\beta$ ranging from $1$ to $5$.
% \begin{figure}
%     \centering
%     \includegraphics[width=7cm]{predpray1.png}
%     \caption{A phase diagram of the predator pray model for $\beta=[1, 2, 3, 4, 5]$, $\alpha = 0.2$,     $\gamma = 0.3$ and  $\delta = 1$}
%     \label{fig:PPM}
% \end{figure}

\begin{figure}
    \centering
    \includegraphics[width=8cm]{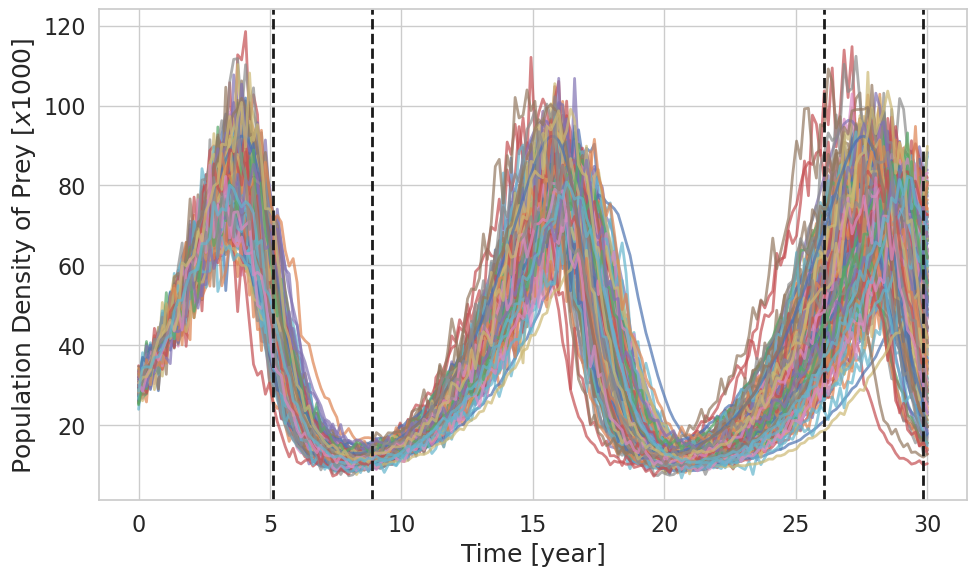}
    \caption{Synthetic prey population data $\bfd = x(t)$ for the Predator-Prey system at noise levels $\sigma = 0\%, 1\%, 2\%, 3\%...10\%$. 200 sampled parameter sets $[\alpha, \beta, \gamma, \delta]$ were used to generate $\bfd$ which are plotted for $t=0$ to $t=30$ years. $200$ equally spaced time points are considered. The four dotted lines indicate an optimal sampling scheme of $sparsity=4$ at $t = 5.1, 8.9, 26.1, 29.9$ years obtained using continuous design variables $\bfw$ (Method 1).}
    \label{fig:PPM}
\end{figure}
In order to estimate the parameters measuring the quantities $x$ or $y$ (or both) are needed.
In practice, it is difficult to measure both species
and therefore we 
assume that the experimentalist is only able to measure the population density of the prey. Using this data the goal is to infer the unknown parameters $\alpha$, $\beta$, $\delta$ and $\gamma$. In practice, it is impossible to obtain a continuous monitoring of $x$ or $y$. Assume that we are given a budget for the measurement process. Similar to the 3-TC model, the experimental
design parameters are the number of measurements and the times that at which they need to be performed.
Finally, in order to learn a Likelihood Free Estimator, we require to have samples from the parameters $\bfq$.
In \cite{LV_param_source}, data on the populations of lynx and hares from the Hudson Bay Company is used to fit the four unknown parameters $\alpha$, $\beta$, $\gamma$ and $\delta$. To exhibit our proposed methods, we assumed lognormal prior distributions with means equal to their fitted parameters and a standard deviation of $5\%$ of the respective means:
\begin{equation}
\label{eq:PPM_lognormal_prior}
\ln(\bfq) \sim \mathcal{N}(\mu_i, \sigma_i^2) \quad \text{for } i = \alpha, \beta, \gamma, \delta
\end{equation}
with numbers specified in Appendix \ref{appendix_A}. Samples are drawn from this prior distribution to generate noisy data samples $\bfd$ (\eqref{forward}) for training. Gaussian noise of $0\%...10\%$ was added to generate the synethetic noisy prey data $\bfd$ (see Figure \ref{fig:PPM}).
\begin{figure}[h]
    \centering
    \includegraphics[width=8cm]{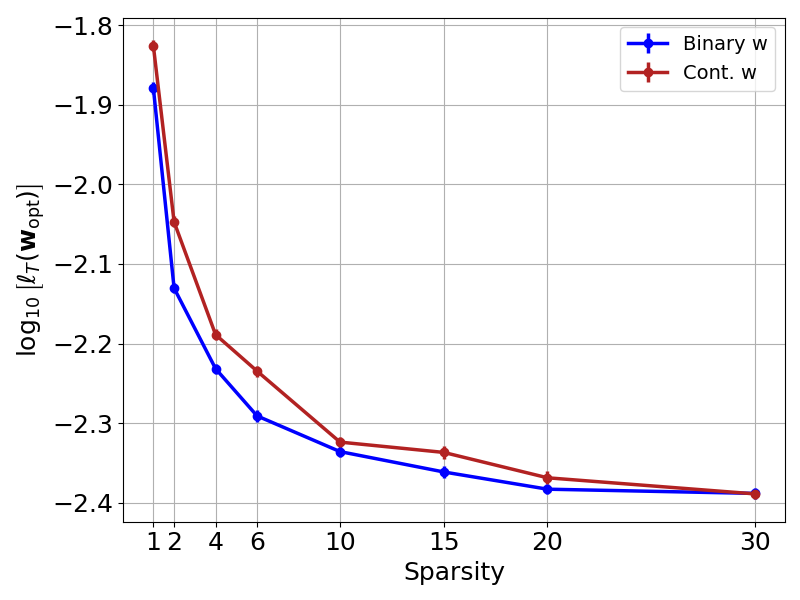}
    \caption{A comparison of the performance of the estimator trained using continuous $\bfw$ (Method 1) and binary $\bfw$ (Method 2) for the Predator-Prey model (PPM). Networks trained for optimal designs $\bfw_{\rm opt}$ at the sparsities shown were evaluated for $\ell_{T}(\bfw)$ over $175K$ unseen samples of $\bfq$ (and corresponding $\bfd$). Error bars that appear point-like are plotted showcasing the Standard Error of the Mean (SEM) for the mean risk $\ell_{T}(\bfw)$ exhibited by each network at its corresponding sparsity. See Appendix \ref{appendix_B} for details.}
    \label{fig:L_curve_PPM}
\end{figure}
\vspace{-5 pt}
\subsection{Numerical Results}
The network structure for the experiments is given in \Cref{alg:net}, trained for continuous $\bfw$ (section \Cref{cont_design_section}) and binary $\bfw$ (\Cref{binary_design_section}). The metric used for the risks $\ell$ correspond to the normalized-Mean-Squared-Error (nMSE) and are described in \Cref{appendix:risk_metric}. Each ODE system was trialled with the two methods. For the 3-TC model, $\bfd$ was generated from the sampled $\bfq$ added with multiplicative Gaussian noise at levels $\sigma = 0\%, 1\%,...,19\%$, and for the PPM model at $\sigma = 0\%, 1\%,...,10\%$. A logarithmically spaced time-grid of size $400$ for $t=0$ to $t=10^{4}$ minutes was used for the 3-TC model as it exhibits a spike at earlier times (see Appendix \ref{appendix_A}). For the PPM system, 200 equally spaced time points were chosen from $t=0$ to $t=30$ years. Here we provide an exposition on the details of the trials.
\begin{table*}[h]
    \caption{\small Total risk $\ell_{T}(\bfw_{\rm opt})$ and parameter risk $\ell_{q}(\bfw_{\rm opt})$ obtained from optimal designs for each ODE system at different sparsities, using both proposed methods. Corresponding averages $\overline{\ell_{T}(\bfw_{\rm rand})}$ and $\overline{\ell_{q}(\bfw_{\rm rand})}$ for 100 networks trained on random $\bfw$ are shown for comparison. See Appendix \ref{appendix_B} for further details.}
    \label{tab:results}
    \centering
    \begin{tabular}{@{}ccccccccc@{}}
        \toprule
        \textbf{ODE} & \textbf{Method} & \textbf{Sparsity} & $\ell_{T}(\bfw_{\text{opt}})$ & $\overline{\ell_{T}(\bfw_{\text{rand}})}$ & $\ell_{q}(\bfw_{\text{opt}})$ & $\overline{\ell_{q}(\bfw_{\text{rand}})}$ \\
        \midrule
        3-TC & Cont. w & 2 & $6.18\times10^{-2}$ & $1.46\times10^{+0}$  & $3.81\times10^{-2}$ &$8.94\times10^{-2}$\\
        3-TC & Cont. w & 6 & $5.42\times10^{-2}$ & $1.87\times10^{-1}$ & $3.84\times10^{-2}$ & $5.79\times10^{-2}$ \\
        3-TC & Cont. w & 10 & $5.29\times10^{-2}$& $8.73\times10^{-2}$ & $3.84\times10^{-2}$ & $4.35\times10^{-2}$ \\
        3-TC & TS (Binary w) & 2 & $5.78\times10^{-2}$ & $8.07\times10^{-2}$ & $3.82\times10^{-2}$ &$3.83\times10^{-2}$ \\
        3-TC & TS (Binary w) & 6 & $5.29\times10^{-2}$ & $5.81\times10^{-2}$ & $3.80\times10^{-2}$ &$3.82\times10^{-2}$ \\
        3-TC & TS (Binary w) & 10 & $5.22\times10^{-2}$ &$5.55\times10^{-2}$ & $3.82\times10^{-2}$ &$3.84\times10^{-2}$ \\
        PPM & Cont. w & 2 & $8.96\times10^{-3}$ & $2.07\times10^{-2}$ & $1.07\times10^{-3}$ & $1.38\times10^{-3}$ \\
        PPM & Cont. w & 4 & $6.45\times10^{-3}$ & $1.43\times10^{-2}$ & $5.45\times10^{-4}$ & $1.11\times10^{-3}$ \\
        PPM & Cont. w & 10 & $4.72\times10^{-3}$ & $9.18\times10^{-3}$ & $4.27\times10^{-4}$ & $5.54\times10^{-4}$ \\
        PPM & TS (Binary w) & 2 & $7.42\times10^{-3}$ & $2.45\times10^{-2}$ & $7.88\times10^{-4}$ & $1.60\times10^{-3}$ \\
        PPM & TS (Binary w) & 4 & $5.87\times10^{-3}$ & $1.19\times10^{-2}$ & $5.73\times10^{-4}$ & $8.18\times10^{-4}$ \\
        PPM & TS (Binary w) & 10 & $4.61\times10^{-3}$ & $8.16\times10^{-3}$ & $3.58\times10^{-4}$ & $6.00\times10^{-4}$ \\
        \bottomrule
    \end{tabular}
\end{table*}
For both training methods, at each iteration $3500$ realizations of $\bfq$ were generated from the prior distributions for both ODE systems. 
Learning rates were chosen by trial and error.
%Through several trials, we observed that the network parameter learning rate $\mu$ and the design vector learning rate $\mu_{\bfw}$ required for convergence was dependent on the ODE system, but not on the training method (Methods 1 and 2) used. 

%For the 3-TC compartment model it was seen that $\mu = 10^{-3} $ and $\mu_{\bfw} = 2 \times10^{-3}$ allowed network training to make reasonable progress towards convergence. Lowering the learning rates further and training for more iterations did not significantly improve the loss. When training with Method 1 in section \ref{cont_design_section}, the first phase of training was to promote sparsity with sparsity parameter $\rho$. With trial and error we found $\rho \in [10^{-4}, 8\times10^{-4}]$ to be optimal with higher values in this range needing to be used to reach lower sparsities. 
We defined sparsity as the number of weights higher than a threshold of $10^{-3}$. For Method 1, the design vector $\bfw$ was initialized as a vector of ones $\bfw_{{\rm init}} = \mathbf{1}$. The training iterations were continued until the desired sparsity level was reached. Weights that were lower than the threshold at the end of the first phase were set to $0$, and a further $5000$ iterations were performed with the obtained $\bfw_{\rm opt}$ fixed, which concluded the training phase. 
%The network for the PPM model was trained similarly, with the same range of $\rho$, but the optimal values for the learning rates that were obtained using trial and error were $\mu = 10^{-4} $ and $\mu_{\bfw} = 2 \times10^{-3}$. The number of iterations used were sufficient for convergence of both models, with convergence being achieved by the 3-TC model at all the sparsities tested in our experiments in $~2000$ iterations, and in $~5000$ iterations for the PPM model. The maximum training time between these models was $\sim8$ hours for each sparsity, a significant improvement over far more time consuming classical experimental design algorithms. Note that to reach smaller sparsities requires more training iterations in the first phase.
For training using Algorithm \ref{alg:sst_binary} (Method 2) at each sparsity, initialization was found to be important for convergence to a $\bfw_{\rm opt}$ after $200$ Tabu iterations. Hyper-parameters such as the size of the subset of neighbours used and the Tabu list length were found using trial and error, and are $10$ and $8$ respectively for both ODEs. For initialization of $\bfw$, a network was trained on a random binary $\bfw$ at the chosen sparsity, and then used as a pre-trained model to train $100$ networks using random binary designs $\bfw_{\rm rand}$ (at fixed sparsity) for a smaller number of iterations. The random $\bfw_{\rm rand}$ that gave the network with the lowest risk $\ell_{T}(\bfw)$ was used to initialize the Tabu Search routine given by Algorithm \ref{alg:sst_binary} to obtain a final $\bfw_{\rm opt}$.
%The total training time for a given sparsity level was found to be $\sim16$ hours using this method. 
The two training methods that we propose as algorithms for fast and efficient experimental design was repeated for each ODE system at a number of different sparsities. For each ODE system, method, and sparsity, the model was evaluated on $175K$ newly sampled $\bfq$'s (and corresponding $\bfd$'s generated as in \eqref{forward}). The results are given in Table \ref{tab:results}. Higher sparsities are expected to give better parameter recoveries and data fits, lowering $\ell_{T}(\bfw)$, while the opposite is expected for lower sparsities. This trend is seen in the comparison of the two methods used in Figure \ref{fig:L_curve_PPM} for the PPM ODE system, and in Figure \ref{fig:L_curve_3TC} for the 3-TC ODE system. Additional experiments are described and included in \Cref{appendix:c}.
\section{Conclusions}
\label{sec:sec6}

In this paper we have introduced a methodology for experimental design for parameter estimation using a Likelihood Free Estimator that involves the use of deep networks. 
Similar to other design methodologies we require the availability of training data, that samples the potential parameters to be recovered. Our methodology is self-supervised. It utilizes parameters to obtain the data under some experimental settings, and then, uses the Likelihood Free Estimator to evaluate the parameters.
Since Likelihood Free Estimators circumvent the need of solving an optimization problem for the parameter, the experimental design problem does not require the solution of a bilevel optimization problem.
We have introduced two design criteria. The first is where the data is multiplied by a continuous weight, and the second, when the data is multiplied by a binary weight. We experiment with these methods on two problems and show that it is possible to obtain an efficient design that can substantially reduce the cost of collecting data that is used in parameter estimation. One natural extension and future direction of the method can be in the problem of sensor placement when considering Partial Differential Equations. However, note that in this case the design space tends to be much larger. It requires further study to test the scaling of our algorithm for such problems.

\section{Acknowledgements}
\label{sec:acknowledgements}
We acknowledge the Canadian Institutes of
Health Research (CIHR) Project Grant PJT-180251. We would like to thank Tamila Kalimullina for help with Figures \ref{fig:fig1} and \ref{fig:3comp}, and Shadab Ahamed for technical support. 
\bibliography{biblio2}

\begin{thebibliography}{44}
\providecommand{\natexlab}[1]{#1}
\providecommand{\url}[1]{\texttt{#1}}
\expandafter\ifx\csname urlstyle\endcsname\relax
  \providecommand{\doi}[1]{doi: #1}\else
  \providecommand{\doi}{doi: \begingroup \urlstyle{rm}\Url}\fi

\bibitem[Allaire et~al.(2001)Allaire, Aubry, and Jouve]{allaubjou}
Allaire, G., Aubry, S., and Jouve, F.
\newblock Eigenfrequency optimization in optimal design.
\newblock \emph{Comp. Meth. Appl. Mech. Eng.}, 190(28):\penalty0 3565--3579,
  2001.

\bibitem[Atkinson \& Donev(1992)Atkinson and Donev]{AtkinsonDonev1992}
Atkinson, A.~C. and Donev, A.~N.
\newblock \emph{Optimum Experimental Designs}.
\newblock Oxford University Press, 1992.

\bibitem[Banga \& Balsa-Canto(2008)Banga and Balsa-Canto]{banga2008parameter}
Banga, J.~R. and Balsa-Canto, E.
\newblock Parameter estimation and optimal experimental design.
\newblock \emph{Essays in biochemistry}, 45:\penalty0 195--210, 2008.

\bibitem[Bardow(2008)]{bardow2008}
Bardow, A.
\newblock Optimal experimental design for ill-posed problems, the meter
  approach.
\newblock \emph{Computers and chemical engineering}, 32, 2008.

\bibitem[Bellman et~al.(1966)Bellman, Kagiwada, and Kalaba]{ecologyInvProblems}
Bellman, R., Kagiwada, H., and Kalaba, R.
\newblock Inverse problems in ecology.
\newblock \emph{J. Theor. Biol.}, 11\penalty0 (1):\penalty0 164--167, 1966.

\bibitem[Bentourkia \& Zaidi(2007)Bentourkia and
  Zaidi]{armanRef2_bentourkia2007tracer}
Bentourkia, M. and Zaidi, H.
\newblock Tracer kinetic modeling in pet.
\newblock \emph{Pet Clinics}, 2\penalty0 (2):\penalty0 267--277, 2007.

\bibitem[Betts(2001)]{BetJ:01}
Betts, J.
\newblock \emph{Practical Methods for Optimal Control using Nonlinear
  Programming}.
\newblock Advances in Design and Control. SIAM, Philadelphia, 2001.

\bibitem[Bj{\"o}rck(1996)]{Bjo96a}
Bj{\"o}rck, {\AA}.
\newblock \emph{Numerical Methods for Least Squares Problems}.
\newblock SIAM, Philadelphia, 1996.

\bibitem[Bock et~al.(2013)Bock, K{\"o}rkel, and
  Schl{\"o}der]{bock2013parameter}
Bock, H.~G., K{\"o}rkel, S., and Schl{\"o}der, J.~P.
\newblock Parameter estimation and optimum experimental design for differential
  equation models.
\newblock \emph{Model Based Parameter Estimation: Theory and Applications},
  pp.\  1--30, 2013.

\bibitem[Boyd \& Vandenberghe(2004)Boyd and Vandenberghe]{boydBook}
Boyd, S. and Vandenberghe, L.
\newblock \emph{Convex optimization}.
\newblock Cambridge University press, 2004.

\bibitem[Bruckstein et~al.(2009)Bruckstein, Donoho, and Elad]{eladReview}
Bruckstein, A., Donoho, D., and Elad, M.
\newblock From sparse solutions of systems of equations to sparse modeling of
  signals and images.
\newblock \emph{SIAM Review}, 51:\penalty0 34--81, 2009.

\bibitem[Chaloner \& Verdinelli(1995)Chaloner and
  Verdinelli]{ChalonerVerdinelli1995}
Chaloner, K. and Verdinelli, I.
\newblock Bayesian experimental design: A review.
\newblock \emph{Statis. Sci.}, 10:\penalty0 237--304, 1995.

\bibitem[Chung \& Haber(2013)Chung and Haber]{ChungHaber12}
Chung, M. and Haber, E.
\newblock Experimental design for biological systems.
\newblock \emph{SIAM Journal on Control and Optimization}, 50:\penalty0
  471--489, 2013.

\bibitem[Didelot et~al.(2011)Didelot, Everitt, Johansen, and
  Lawson]{didelot2011likelihood}
Didelot, X., Everitt, R.~G., Johansen, A.~M., and Lawson, D.~J.
\newblock Likelihood-free estimation of model evidence.
\newblock 2011.

\bibitem[Donoho(2006)]{donoho04sparsest}
Donoho, D.~D.
\newblock For most large underdetermined systems of linear equations the
  minimal $\ell_1$-norm solution is also the sparsest solution.
\newblock \emph{Communications on Pure and Applied Mathematics}, 59\penalty0
  (6):\penalty0 797--829, 2006.

\bibitem[Ette \& Williams(2007)Ette and Williams]{pharmacometricsBook}
Ette, E.~I. and Williams, P. (eds.).
\newblock \emph{Pharmacometrics: The Science of Quantitative Pharmacology}.
\newblock Wiley-Interscience, New York, 2007.

\bibitem[Feng et~al.(1994)Feng, Wang, and Yan]{Feng_Input_Function_source}
Feng, D., Wang, X., and Yan, H.
\newblock A computer simulation study on the input function sampling schedules
  in tracer kinetic modeling with positron emission tomography (pet).
\newblock \emph{Computer Methods and Programs in Biomedicine}, 45\penalty0
  (3):\penalty0 175--186, 1994.
\newblock ISSN 0169-2607.
\newblock \doi{https://doi.org/10.1016/0169-2607(94)90201-1}.
\newblock URL
  \url{https://www.sciencedirect.com/science/article/pii/0169260794902011}.

\bibitem[Gendreau \& Potvin(2005)Gendreau and Potvin]{gendreau2005tabu}
Gendreau, M. and Potvin, J.-Y.
\newblock Tabu search.
\newblock \emph{Search methodologies: introductory tutorials in optimization
  and decision support techniques}, pp.\  165--186, 2005.

\bibitem[Haber et~al.(2008)Haber, Horesh, and Tenorio]{HaberHoreshTenorio08}
Haber, E., Horesh, L., and Tenorio, L.
\newblock Numerical methods for experimental design of large-scale linear
  ill-posed inverse problems.
\newblock \emph{Inverse Problems}, 24, 2008.

\bibitem[Haber et~al.(2009)Haber, Horesh, and Tenorio]{HaberHoreshTenorio09}
Haber, E., Horesh, L., and Tenorio, L.
\newblock Numerical methods for experimental design of nonlinear ill-posed
  inverse problems.
\newblock \emph{Inverse Problems}, 25, 2009.

\bibitem[Helms(2008)]{PrincCompBio}
Helms, V. (ed.).
\newblock \emph{Principles of Computational Cell Biology: From Protein
  Complexes to Cellular Networks}.
\newblock Wiley-VCH, New York, 2008.

\bibitem[Hertz et~al.(1995)Hertz, Taillard, and De~Werra]{hertz1995tutorial}
Hertz, A., Taillard, E., and De~Werra, D.
\newblock A tutorial on tabu search.
\newblock In \emph{Proc. of Giornate di Lavoro AIRO}, volume~95, pp.\  13--24,
  1995.

\bibitem[Hunt et~al.(2007)Hunt, Kostelich, and Szunyogh]{hun07a}
Hunt, B.~R., Kostelich, E.~J., and Szunyogh, I.
\newblock Efficient data assimilation for spatiotemporal chaos: A local
  ensemble transform {K}alman filter.
\newblock \emph{Physica D}, 230:\penalty0 112--126, 2007.

\bibitem[Jaiswal et~al.(2020)Jaiswal, Babu, Zadeh, Banerjee, and
  Makedon]{jaiswal2020survey}
Jaiswal, A., Babu, A.~R., Zadeh, M.~Z., Banerjee, D., and Makedon, F.
\newblock A survey on contrastive self-supervised learning.
\newblock \emph{Technologies}, 9\penalty0 (1):\penalty0 2, 2020.

\bibitem[Keane et~al.(2017)Keane, Krauskopf, and
  Postlethwaite]{keane2017climate}
Keane, A., Krauskopf, B., and Postlethwaite, C.~M.
\newblock Climate models with delay differential equations.
\newblock \emph{Chaos: An Interdisciplinary Journal of Nonlinear Science},
  27\penalty0 (11), 2017.

\bibitem[Kletting et~al.(2016)Kletting, Schuchardt, Kulkarni, Shahinfar, Singh,
  Glatting, Baum, and Beer]{KlettingPaper}
Kletting, P., Schuchardt, C., Kulkarni, H., Shahinfar, M., Singh, A., Glatting,
  G., Baum, R.~P., and Beer, A.
\newblock Investigating the effect of ligand amount and injected therapeutic
  activity: A simulation study for 177lu-labeled psma-targeting peptides.
\newblock \emph{PloS one}, 11:\penalty0 e0162303, 09 2016.
\newblock \doi{10.1371/journal.pone.0162303}.

\bibitem[Lenhart \& Workman(2007)Lenhart and Workman]{optcontrolBook}
Lenhart, S. and Workman, J. (eds.).
\newblock \emph{Optimal Control Applied to Biological Models (1st ed.)}.
\newblock Chapman and Hall/CRC, New York, 2007.

\bibitem[Li et~al.(2018)Li, Yu, Shahabi, and Liu]{li2018diffusion}
Li, Y., Yu, R., Shahabi, C., and Liu, Y.
\newblock {Diffusion Convolutional Recurrent Neural Network: Data-Driven
  Traffic Forecasting}.
\newblock In \emph{International Conference on Learning Representations}, 2018.

\bibitem[Mahaffy(2009)]{LV_param_source}
Mahaffy, J.
\newblock Qualitative analysis of 2{D} systems of {O}rdinary {D}ifferential
  {E}quations.
\newblock
  \url{https://jmahaffy.sdsu.edu/courses/f09/math636/lectures/lotka/qualde2.html},
  2009.
\newblock Accessed: Dec 25, 2023.

\bibitem[Matthew Sainsbury-Dale \& Huser(2024)Matthew Sainsbury-Dale and
  Huser]{sainsbury2022fast}
Matthew Sainsbury-Dale, A. Z.-M. and Huser, R.
\newblock Likelihood-free parameter estimation with neural bayes estimators.
\newblock \emph{The American Statistician}, 78\penalty0 (1):\penalty0 1--14,
  2024.
\newblock \doi{10.1080/00031305.2023.2249522}.
\newblock URL \url{https://doi.org/10.1080/00031305.2023.2249522}.

\bibitem[McLellan(1994)]{mclellan}
McLellan, P.
\newblock A differential-algebraic perspective on nonlinear controller design
  methodologies.
\newblock \emph{Chem. Eng. Science}, 49:\penalty0 1663--1679, 1994.

\bibitem[Morris et~al.(2004)Morris, Endres, Schmidt, Christian, Muzic, and
  Fisher]{armanRef1_morris2004kinetic}
Morris, E.~D., Endres, C.~J., Schmidt, K.~C., Christian, B.~T., Muzic, R.~F.,
  and Fisher, R.~E.
\newblock Kinetic modeling in positron emission tomography.
\newblock \emph{Emission Tomography: The Fundamentals of PET and SPECT},
  46\penalty0 (1):\penalty0 499--540, 2004.

\bibitem[Papamakarios \& Murray(2016)Papamakarios and
  Murray]{papamakarios2019neural_3}
Papamakarios, G. and Murray, I.
\newblock Fast $\epsilon$-free inference of simulation models with bayesian
  conditional density estimation.
\newblock In \emph{Proceedings of the 30th International Conference on Neural
  Information Processing Systems}, NIPS'16, pp.\  1036–1044, Red Hook, NY,
  USA, 2016. Curran Associates Inc.
\newblock ISBN 9781510838819.

\bibitem[Papamakarios et~al.(2017)Papamakarios, Pavlakou, and
  Murray]{papamakarios2019neural_1}
Papamakarios, G., Pavlakou, T., and Murray, I.
\newblock Masked autoregressive flow for density estimation.
\newblock In \emph{Proceedings of the 31st International Conference on Neural
  Information Processing Systems}, NIPS'17, pp.\  2335–2344, Red Hook, NY,
  USA, 2017. Curran Associates Inc.
\newblock ISBN 9781510860964.

\bibitem[Papamakarios et~al.(2019)Papamakarios, Sterratt, and
  Murray]{papamakarios2019neural_2}
Papamakarios, G., Sterratt, D., and Murray, I.
\newblock Sequential neural likelihood: Fast likelihood-free inference with
  autoregressive flows.
\newblock In Chaudhuri, K. and Sugiyama, M. (eds.), \emph{Proceedings of the
  Twenty-Second International Conference on Artificial Intelligence and
  Statistics}, volume~89 of \emph{Proceedings of Machine Learning Research},
  pp.\  837--848. PMLR, 16--18 Apr 2019.
\newblock URL \url{https://proceedings.mlr.press/v89/papamakarios19a.html}.

\bibitem[Pukelsheim(1993)]{Pukelsheim93}
Pukelsheim, F.
\newblock \emph{Optimal design of experiments}.
\newblock John Wiley \& Sons, 1993.

\bibitem[Ushijima \& Yeh(2015)Ushijima and Yeh]{ushijima2015experimental}
Ushijima, T.~T. and Yeh, W.~W.
\newblock Experimental design for estimating unknown hydraulic conductivity in
  an aquifer using a genetic algorithm and reduced order model.
\newblock \emph{Advances in Water Resources}, 86:\penalty0 193--208, 2015.

\bibitem[Wagner(1969)]{CompModelRef3}
Wagner, J.~G.
\newblock Pharmacokinetics: 10. introduction to compartment models.
\newblock \emph{Drug Intelligence \& Clinical Pharmacy}, 3\penalty0
  (9):\penalty0 250--257, 1969.
\newblock \doi{10.1177/106002806900300904}.
\newblock URL \url{https://doi.org/10.1177/106002806900300904}.

\bibitem[Wang et~al.(2023)Wang, Martins, and Du]{wang2023optimal}
Wang, L., Martins, J.~R., and Du, X.
\newblock Optimal experimental design-based reduced order modeling for learning
  optimal aerodynamic designs.
\newblock In \emph{AIAA AVIATION 2023 Forum}, pp.\  3716, 2023.

\bibitem[Yang et~al.(2023)Yang, Zhang, Song, Hong, Xu, Zhao, Zhang, Cui, and
  Yang]{yang2023diffusion}
Yang, L., Zhang, Z., Song, Y., Hong, S., Xu, R., Zhao, Y., Zhang, W., Cui, B.,
  and Yang, M.-H.
\newblock Diffusion models: A comprehensive survey of methods and applications.
\newblock \emph{ACM Computing Surveys}, 56\penalty0 (4):\penalty0 1--39, 2023.

\bibitem[Zakariaei et~al.(2023)Zakariaei, Paranj, Abdollahi, and
  Rahmim]{ZakariaeiP1449}
Zakariaei, N., Paranj, A.~F., Abdollahi, H., and Rahmim, A.
\newblock Using the cluster gauss newton algorithm to estimate theranostic
  pharmacokinetic model parameters.
\newblock \emph{Journal of Nuclear Medicine}, 64\penalty0 (supplement
  1):\penalty0 P1449--P1449, 2023.
\newblock ISSN 0161-5505.
\newblock URL \url{https://jnm.snmjournals.org/content/64/supplement_1/P1449}.

\bibitem[Zenker et~al.(2007)Zenker, Rubin, and Clermont]{physiologyInvProblems}
Zenker, S., Rubin, J., and Clermont, G.
\newblock From inverse problems in mathematical physiology to quantitative
  differential diagnoses.
\newblock \emph{PLOS Computational Biology}, 3\penalty0 (11):\penalty0 1--15,
  2007.

\bibitem[Zhai et~al.(2019)Zhai, Oliver, Kolesnikov, and Beyer]{zhai2019s4l}
Zhai, X., Oliver, A., Kolesnikov, A., and Beyer, L.
\newblock S4l: Self-supervised semi-supervised learning.
\newblock In \emph{Proceedings of the IEEE/CVF international conference on
  computer vision}, pp.\  1476--1485, 2019.

\bibitem[Zhang et~al.(2022)Zhang, Feng, Gong, Lee, Lomonaco, and
  Zhao]{CompModelRef2}
Zhang, P., Feng, K., Gong, Y., Lee, J., Lomonaco, S., and Zhao, L.
\newblock Usage of compartmental models in predicting covid-19 outbreaks.
\newblock \emph{The AAPS Journal}, 24\penalty0 (5):\penalty0 98, 2022.
\newblock ISSN 1550-7416.
\newblock \doi{10.1208/s12248-022-00743-9}.
\newblock URL \url{https://doi.org/10.1208/s12248-022-00743-9}.
\newblock PMID: 36056223.

\end{thebibliography}
\bibliographystyle{icml2024}

%%%%%%%%%%%%%%%%%%%%%%%%%%%%%%%%%%%%%%%%%%%%%%%%%%%%%%%%%%%%%%%%%%%%%%%%%%%%%%%
%%%%%%%%%%%%%%%%%%%%%%%%%%%%%%%%%%%%%%%%%%%%%%%%%%%%%%%%%%%%%%%%%%%%%%%%%%%%%%%
% APPENDIX
%%%%%%%%%%%%%%%%%%%%%%%%%%%%%%%%%%%%%%%%%%%%%%%%%%%%%%%%%%%%%%%%%%%%%%%%%%%%%%%
%%%%%%%%%%%%%%%%%%%%%%%%%%%%%%%%%%%%%%%%%%%%%%%%%%%%%%%%%%%%%%%%%%%%%%%%%%%%%%%
\newpage
\appendix
\onecolumn

\section{Appendix: Additional Details on ODE Models and Setup}
\label{appendix_A}
Here we provide additional details on our experiments. For all experiments concerned, the network structure given by Algorithm \ref{alg:net} used $nlayers=3$, and no layer normalization was used. The $SiLU$ function was used as the activation function. Further results of our experimental runs are given in the tables shown herein. 

The samples $\bfq$ (and therefore $\bfd$ that were generated from them) were generated from a lognormal prior distribution for each of the parameters in $\bfq$ as mentioned in the main text (\eqref{eq:3TC_logprior} and \eqref{eq:PPM_lognormal_prior}). The mean $\mu$ and standard deviation $\sigma$ parameters of a lognormal distribution are related to the mean $\mu_{q}$ and standard deviation $\sigma_{q}$ of the random variable $q$ as follows:
\begin{equation}
\label{eq:lognormal_formula}
\mu = \ln \left( \frac{\mu_q^2}{\sqrt{\mu_q^2 + \sigma_q^2}} \right), \quad \sigma^2 = \ln \left(1 + \frac{\sigma_q^2}{\mu_q^2}\right)
\end{equation}
where $q$ in our case refers to each of the parameters being used in the corresponding ODE model. 
\subsection{Model Setup: 3-Tissue Compartment Model}
\begin{figure}[h]
    \centering
    \includegraphics[width=8cm]{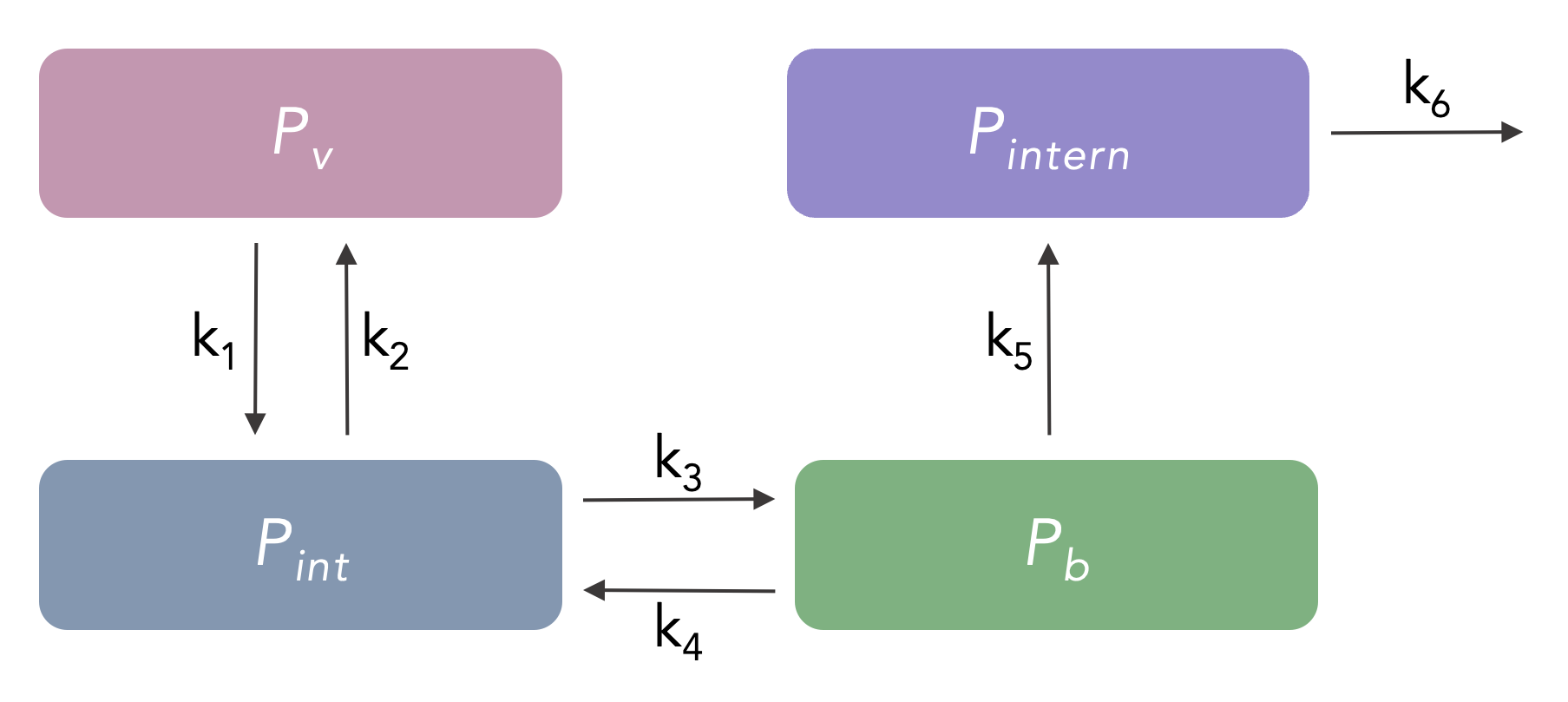}
    % \vspace{-10 pt}  % Adjust the length as needed
    \caption{The three-tissue compartment model that describes radiopharmaceutical dynamics, containing six different parameters to be estimated.
    The compartment $P_v$ is a known source term. A PET scan allows the noisy measurement of $P_{int}+P_{b}+P_{intern}$.}
    \label{fig:3comp}
\end{figure}

A diagram of the 4 compartments in the 3-TC ODE model is provided in Figure \ref{fig:3comp}. The source function $P_{v}$ is a known blood input function (prior to entry into tissue).  
% \vspace{-8 pt}  % Adjust the length as needed
In our experiments for the ODE system, we used the input function proposed in \cite{Feng_Input_Function_source} given by:
\begin{equation}
\begin{aligned}
    \label{Fenginput}
    P_{v}(t) = &(A_{1}t - A_{2} - A_{3})\exp(-\lambda_{1}t) \\
    &+A_{2}\exp(-\lambda_{2}t)+A_{3}\exp(-\lambda_{3}t)
\end{aligned}
\end{equation}
We fit \eqref{Fenginput} to patient data available to us at our institution, arriving at $A_{1} = 408.87, A_{2}=A_{3}=14.78$ and  $\lambda_{1}=-8.46, \lambda_{2}= \lambda_{3}=-0.1362$ which we used in \eqref{Fenginput} to obtain our input function $P_{v}$. This input function exhibits a sharp spike in concentration at earlier times, and as such, we selected a log-spaced discretization of $400$ points for $\bft$ from $t=0$ to $t=10^{4}$ minutes. For the initial conditions we set $P_{v}$, $P_{int}$, $P_{b}$ and $P_{intern}$ to $0$ $\rm kBq/ml$.

The means and standard deviations $\mu_{q}$ and $\sigma_{q}$ for the prior distribution of the parameters $\mu_{i}, i \in\{k_{1},k_{2},k_{4},k_{5},k_{6}\}$, and $\sigma_{i}, i \in\{k_{1},k_{2},k_{4},k_{5},k_{6}\}$ that were used were $\mu_{k_{1}}=1.5\times10^{-2}$, $\mu_{k_{2}}=1.6\times10^{-3}$, $\mu_{k_{3}}=121.0$, $\mu_{k_{4}}=4\times10^{-2}$, $\mu_{k_{5}}=1\times10^{-3}$, $\mu_{k_{6}}=2\times10^{-4}$, and  $\sigma_{k_{1}}=0.2\mu_{k_{1}}, \sigma_{k_{2}} = 0.2\mu_{k_{2}}, \sigma_{k_{3}} =0.2\mu_{k_{3}}, \sigma_{k_{4}}=0.2\mu_{k_{4}},    \sigma_{k_{5}}=0.2\mu_{k_{5}}, \sigma_{k_{6}}=0.2\mu_{k_{6}}$ respectively. As can be seen, we took the standard deviations of the parameters to be $20\%$ of the means. The noise $\sigma$ added to generated $\bfd$ were randomly picked at each sampling from $\{0\%, 1\%, 2\%...19\%\}$.
%%%%%%%%%%%%%%%%%%%%%%%%%%%%%%%%%%%%%%%%%%%%%%%%%%%%%%%%%%%%%%%%%%%%%%%%%%%%%%%
%%%%%%%%%%%%%%%%%%%%%%%%%%%%%%%%%%%%%%%%%%%%%%%%%%%%%%%%%%%%%%%%%%%%%%%%%%%%%%%
\subsection{2: PPM Model Setup}
For the Predator-Prey Model described in Section \ref{sec:sec5}, we used an equally spaced time grid of 200 points from $t=0$ to $t=30$ years. In \cite{LV_param_source}, data on the populations of lynx and hares from the Hudson Bay Company is used to fit the four unknown parameters $\alpha$, $\beta$, $\gamma$ and $\delta$, where they used an initial population of hares (prey) and lynx (predator) as 30 and 4 (in units of thousands) respectively, which we used. We took the fitted parameters obtained by \cite{LV_param_source} as the means of the respective prior distributions of the parameters $k_{1}, k_{2}, k_{3}, k_{4}, k_{5}$ and $k_{6}$, and $5\%$ of the means as their respective standard deviations. The means and standard deviations of the parameters for their respective prior distributions were: $(\mu_{\alpha}, \sigma_{\alpha}) = (0.4, 0.05\times0.4)$, $(\mu_{\beta}, \sigma_{\beta}) = (0.018,0.05\times0.018)$, $(\mu_{\gamma}, \sigma_{\gamma}) = (0.8, 0.05\times0.8)$, $(\mu_{\delta}, \sigma_{\delta}) = (0.023, 0.05\times0.023)$. The noise $\sigma$ added to the generated $\bfd$ was randomly picked at each sampling from $\{0\%, 1\%, 2\%...10\%\}$. 
\subsection{Risk Metric}
\label{appendix:risk_metric}
The loss function \eqref{totalrisk} is a sum of the data risk $\ell_{d}(\omega)$ and the parameter risk $\ell_{q}(\omega)$. For the data-risk $\ell_{d}(\omega)$ we used the trapezoidal rule for variable time widths (owing to our use of logarithmically spaced time points). Note that the risks $\ell(\omega)$ reported in our results and that used during training had the scaling $k$ shown below in \eqref{eq:data_risk_scaling}, and whose discretized form was used during training.
\begin{equation}
\begin{aligned}
\label{eq:data_risk_scaling}
\ell_{d}(\omega) &= {\mathbb E}_{\bfq, \bfepsilon}\, \hf \int_0^T  k \cdot \|F(\widehat \bfq_i, \bfomega) - \bfd_i(t) \|^2 dt \\
k &= \frac{1}{\frac{1}{2}\int_{0}^{T} \|\bfd_i(t)\|^{2} dt}
\end{aligned}
\end{equation}
where $k$ is a scaling factor. Similarly for the parameter recovery: risk $\ell_{q}(\omega)$:
\begin{eqnarray}
    \label{param_risk_scaling}
    \ell_q(\bfomega) &=& 
    {\mathbb E}_{\bfq, \bfepsilon}\, \hf \cdot k \cdot \|\widehat \bfq - \bfq \|^2 \\
    k &=& \frac{1}{ \frac{1}{2}\|\bfq\|^{2}}
\end{eqnarray}

Thus the metric for the parameter risk $\ell_q(\bfw)$, the data risk $\ell_d(\bfw)$ and the total risk $\ell_T(\bfw) = \ell_q(\bfw) + \ell_d(\bfw)$ that are reported in all our experiments, with the exception of \Cref{table:A_opt_performance_comparison} where the mean-squared-error (MSE) is shown, correspond to the normalized mean-squared-error (nMSE).
% \onecolumn 
% \newpage
\section{Appendix: Numerical Details of Figures and Tables}
\label{appendix_B}
% \subsection{Numerical details: Figures and Tables}

In Figure \ref{fig:L_curve_3TC} and Figure \ref{fig:L_curve_PPM} we exhibit the performance of networks trained using the proposed method at different sparsities. For each training method at each sparsity, a new batch of $175K$ samples of $\bfq$ (and corresponding noisy $\bfd$) were sampled for the trained optimal network to be evaluated on. For the computation of $\ell_{T}(\bfw)$, the network was evaluated on $50$ sets of $3500$ newly sampled $q$ and $d$, amounting to an evaluation over $175K$ unseen samples. The mean value for the total risk $\ell_{T}(\bfw)$ of the optimal network over these $50$ sets was computed and taken as $\ell_{T}(\bfw_{\rm opt})$, and the Standard Error of the Mean (SEM) for this value was calculated. The $\ell_{T}(\bfw_{\rm opt})$ are plotted against $sparsity$ (number of non-zero elements in $\bfw$) in Figure \ref{fig:L_curve_3TC} and in Figure \ref{fig:L_curve_PPM}. The corresponding numerical values are presented in Table \ref{tab:appendix_totalRisk_SEM}. Note that in Figure \ref{fig:L_curve_3TC} and Figure \ref{fig:L_curve_PPM}, the error bars look point-like due to the SEM($\ell_{T}(\bfw)$) values being $\sim 10^{-2}$ smaller than the values of $\ell_{T}(\bfw)$ that are plotted (see Table \ref{tab:appendix_totalRisk_SEM}). The error bars in the figures were calculated as:
\begin{eqnarray}
    \label{eq: appendix_L_curve_error_bar_calc}
    \log_{10}\left(\ell_{T}(\bfw) + \text{SEM}(\ell_{T}(\bfw)\right) - \log_{10}\left(\ell_{T}(\bfw)\right)
\end{eqnarray}
The corresponding numerical values for $\ell_{T}(\bfw_{\rm opt})$ and SEM($\ell_{T}(\bfw_{\rm opt})$) are in Table \ref{tab:appendix_totalRisk_SEM}. 

We chose to conduct numerical experiments to compare the risks obtained by our methods to random designs for a subset of sparsities. For the 3-TC model, this was at $sparsity = 2,6,10,20$, and at $sparsity = 2,4,10,20$ for the PPM system of equations.

Namely, the risks $\ell_{T}(\bfw)$, $\ell_{q}(\bfw)$ obtained by training optimal networks for a given method and sparsity are compared against networks trained on random $\bfw$. In Table \ref{tab:results}, Table \ref{tab:appendix_risk_results} and Table \ref{tab:Appendix:ParameterRecoveryresults}, for each ODE model, method, and sparsity, 100 networks were trained on random designs $\bfw_{rand}$ for comparison. For the random runs for continuous $\bfw$, the entries of $\bfw_{\rm rand}$ were sampled from a uniform distribution ${\bfw}_{\rm{rand}} \sim \text{Uniform}(0, 2)$. For binary design variables, $\bfw_{rand}$ had a random subset of weights (equal to $sparsity$) set to $1$, with the rest set to $0$.
For the computation of each $\ell_{T}(\bfw_{\rm opt})$, the optimal network was evaluated on $50$ sets of $3500$ newly sampled $\bfq$ and $\bfd$, amounting to an evaluation over $175K$ unseen samples. A mean value for the risk of the optimal network over these samples was computed and taken as $\ell_{T}(\bfw)_{\rm opt}$, and the Standard Error of the Mean for this value was calculated. This is given as the {\bf SEM}($\ell_{T}(\bfw)_{\rm opt}$) column in Table \ref{tab:appendix_risk_results} and Table \ref{tab:appendix_totalRisk_SEM}. The parameter risks $\ell_{q}(\bfw_{\rm opt})$ for the optimal networks were calculated similarly. Each of the $100$ networks trained on random designs $\bfw_{\rm rand}$ were evaluated for $\ell_{T}(\bfw_{\rm rand})$ on $70K$ new samples of $\bfq$ and $\bfd$. The {\bf STD}($\ell_{T}(\bfw_{\rm rand})$) and {\bf STD}($\ell_{q}(\bfw_{\rm rand}))$ columns in Table \ref{tab:appendix_risk_results} and Table \ref{tab:Appendix:ParameterRecoveryresults} are the standard deviations of the corresponding risks among the 100 networks trained on random $\bfw_{rand}$ for each corresponding entry in the tables. In Table \ref{tab:ComparingWithMins} we confirm that the total risks for the trained optimal networks $\ell_{T}(\bfw_{\rm opt})$ are lower than the risk obtained by the best performing network trained on random $\bfw_{\rm rand}$ for each entry. 
\begin{table*}[t]
    \caption{\small Total risks $\ell_{T}(\bfw_{\rm opt})$ obtained by networks trained for optimal designs $\bfw_{\rm opt}$ on $175K$ unseen samples of $\bfq$ and $\bfd$. The Standard Error of the Mean (SEM) for each are also presented.}
    \label{tab:appendix_totalRisk_SEM}
    \centering
    \begin{tabular}{@{}ccccc@{}}
        \toprule
        \textbf{ODE} & \textbf{Method} & \textbf{Sparsity} & $\ell_{T}(\bfw_{\text{opt}})$ & \textbf{SEM}($\ell_{T}(\bfw_{\text{opt}})$) \\
        \midrule
        3-TC & Cont. w & 1 & $6.44\times10^{-2}$ & $1.00\times10^{-3}$ \\
        3-TC & Cont. w & 2 & $6.18\times10^{-2}$ & $1.28\times10^{-3}$ \\
        3-TC & Cont. w & 4 & $5.48\times10^{-2}$ & $1.18\times10^{-3}$ \\
        3-TC & Cont. w & 6 & $5.42\times10^{-2}$ & $1.17\times10^{-3}$ \\
        3-TC & Cont. w & 10 & $5.29\times10^{-2}$ & $1.04\times10^{-3}$ \\
        3-TC & Cont. w & 15 & $5.17\times10^{-2}$ & $8.70\times10^{-4}$ \\
        3-TC & Cont. w & 20 & $5.125\times10^{-2}$ & $7.57\times10^{-4}$ \\
        3-TC & Cont. w & 30 & $5.120\times10^{-2}$ & $1.13\times10^{-3}$ \\
        3-TC & TS (Binary w) & 1 & $6.28\times10^{-2}$ & $1.01\times10^{-3}$ \\
        3-TC & TS (Binary w) & 2 & $5.78\times10^{-2}$ & $1.25\times10^{-3}$ \\
        3-TC & TS (Binary w) & 4 & $5.48\times10^{-2}$ & $1.07\times10^{-3}$ \\
        3-TC & TS (Binary w) & 6 & $5.29\times10^{-2}$ & $1.07\times10^{-3}$ \\
        3-TC & TS (Binary w) & 10 & $5.22\times10^{-2}$ & $1.05\times10^{-3}$ \\
        3-TC & TS (Binary w) & 15 & $5.17\times10^{-2}$ & $8.41\times10^{-4}$ \\
        3-TC & TS (Binary w) & 20 & $5.14\times10^{-2}$ & $8.63\times10^{-4}$ \\
        3-TC & TS (Binary w) & 30 & $5.11\times10^{-2}$ & $1.10\times10^{-3}$ \\
        PPM & Cont. w & 1 & $1.49\times10^{-2}$ & $2.32\times10^{-4}$ \\
        PPM & Cont. w & 2 & $8.96\times10^{-3}$ & $1.33\times10^{-4}$ \\
        PPM & Cont. w & 4 & $6.45\times10^{-3}$ & $1.23\times10^{-4}$ \\
        PPM & Cont. w & 6 & $5.83\times10^{-3}$ & $9.33\times10^{-5}$ \\
        PPM & Cont. w & 10 & $4.72\times10^{-3}$ & $6.90\times10^{-5}$ \\
        PPM & Cont. w & 15 & $4.60\times10^{-3}$ & $9.12\times10^{-5}$ \\
        PPM & Cont. w & 20 & $4.28\times10^{-3}$ & $8.06\times10^{-5}$ \\
        PPM & Cont. w & 30 & $4.08\times10^{-3}$ & $5.83\times10^{-5}$ \\
        PPM & TS (Binary w) & 1 & $1.32\times10^{-2}$ & $2.33\times10^{-4}$ \\
        PPM & TS (Binary w) & 2 & $7.42\times10^{-3}$ & $8.27\times10^{-5}$ \\
        PPM & TS (Binary w) & 4 & $5.87\times10^{-3}$ & $8.77\times10^{-5}$ \\
        PPM & TS (Binary w) & 6 & $5.12\times10^{-3}$ & $8.84\times10^{-5}$ \\
        PPM & TS (Binary w) & 10 & $4.61\times10^{-3}$ & $6.68\times10^{-5}$ \\
        PPM & TS (Binary w) & 15 & $4.35\times10^{-3}$ & $7.85\times10^{-5}$ \\
        PPM & TS (Binary w) & 20 & $4.15\times10^{-3}$ & $4.88\times10^{-5}$ \\
        PPM & TS (Binary w) & 30 & $4.09\times10^{-3}$ & $6.44\times10^{-5}$ \\
        \bottomrule
    \end{tabular}
\end{table*}

%%%

\begin{table*}[t]
    \caption{\small Experimental results for total risks $\ell_{T}(\bfw)$. The $\ell_{T}(\bfw)$ were computed over $175k$ unseen samples for each ODE system at the given sparsity values. The corresponding Standard Error of the Mean (SEM) are shown. Also given are the mean parameter risks $\overline{\ell_{T}(\bfw_{\rm rand})}$ obtained by 100 networks trained on random designs $\bfw_{\rm rand}$. The standard deviations of $\ell_{T}(\bfw_{\rm rand})$ and the percentage difference of $\ell_{T}(\bfw_{\rm opt})$ from the random runs are similarly presented. $\bf{SP}$ refers to the sparsity of the corresponding row.}
    \label{tab:appendix_risk_results}
    \centering
    \begin{tabular}{@{}cccccccc@{}}
        \toprule
        \textbf{ODE} & \textbf{Method} & \textbf{SP} & $\ell_{T}(\bfw_{\rm{opt}})$ & \textbf{SEM}($\ell_{T}(\bfw_{\rm{opt}})$) & $\overline{\ell_{T}(\bfw_{\rm{rand}})}$ & $\frac{\ell_{T}(\bfw_{\rm{opt})} - \overline{\ell_{T}(\bfw_{\rm{rand}})}}{\overline{\ell_{T}(\bfw_{\rm{rand}})}} \times 100\%$ & $\textbf{STD}(\ell_{T}(\bfw_{\rm rand}))$ \\
        \midrule
        3-TC & Cont. w & 2& $6.18\times10^{-2}$ &$1.28 \times 10^{-3}$ &$1.46\times10^{+0}$ &$-95.78\%$ &$2.05\times10^{+0}$ \\
        3-TC & Cont. w & 6 & $5.42\times10^{-2}$ &$1.17 \times 10^{-3}$ &$1.87\times10^{-1}$ &$-71.10\%$ &$2.88\times10^{-1}$ \\
        3-TC & Cont. w & 10 & $5.29\times10^{-2}$ &$1.04 \times 10^{-3}$ & $8.73\times10^{-2}$&$-39.43\%$ &$1.49\times10^{-1}$ \\
        3-TC & Cont. w & 20 & $5.125\times10^{-2}$ & $8.70 \times 10^{-4}$& $5.98\times10^{-2}$& $-14.30\%$& $2.44\times10^{-3}$\\
        3-TC & TS (Binary w) & 2 &$5.78\times10^{-2}$ &$1.25 \times 10^{-3}$ &$8.07\times10^{-2}$ &$-28.37\%$ &$8.69\times10^{-3}$ \\
        3-TC & TS (Binary w) & 6 &$5.29\times10^{-2}$ &$1.07 \times 10^{-3}$ &$5.81\times10^{-2}$ &$-9.06\%$ &$2.19\times10^{-3}$ \\
        3-TC & TS (Binary w) & 10 &$5.22\times10^{-2}$ &$1.05 \times 10^{-3}$ &$5.55\times10^{-2}$ &$-5.92\%$ &$1.43\times10^{-3}$ \\
        3-TC & TS (Binary w) & 20 & $5.14\times10^{-2}$&$8.63 \times 10^{-4}$ &$5.35\times10^{-2}$ &$-3.76\%$ & $7.66\times10^{-4}$ \\
        PPM & Cont. w & 2 & $8.96\times10^{-3}$ & $1.33 \times 10^{-4}$& $2.07\times10^{-2}$&$-56.72\%$ & $7.46\times10^{-3}$\\
        PPM & Cont. w & 4 & $6.45\times10^{-3}$ & $1.23 \times 10^{-4}$& $1.43\times10^{-2}$&$-54.89\%$ & $8.62\times10^{-3}$\\
        PPM & Cont. w & 10 & $4.72\times10^{-3}$ &$6.90 \times 10^{-5}$ & $9.18\times10^{-3}$& $-48.59\%$&$2.33\times10^{-3}$ \\
        PPM & Cont. w & 20 & $4.28\times10^{-3}$ &$8.06 \times 10^{-5}$ & $7.16\times10^{-3}$& $-40.22\%$& $3.14\times10^{-3}$\\
        PPM & TS (Binary w) & 2 & $7.42 \times 10^{-3}$ & $8.27 \times 10^{-5}$ & $2.45 \times 10^{-2}$ & $-69.77\%$ & $8.93\times10^{-3}$\\
        PPM & TS (Binary w) & 4 & $5.87 \times 10^{-3}$ & $8.77 \times 10^{-5}$ & $1.19 \times 10^{-2}$ & $-50.89\%$ & $4.76\times10^{-3}$\\
        PPM & TS (Binary w) & 10 & $4.61 \times 10^{-3}$ &$6.68 \times 10^{-5}$  & $8.16 \times 10^{-3}$ & $-43.51\%$&$5.82\times10^{-3}$ \\
        PPM & TS (Binary w) & 20 & $4.15 \times 10^{-3}$ &$4.88 \times 10^{-5}$ & $5.63 \times 10^{-3}$ &$-26.25\%$ & $2.16\times10^{-3}$  \\
        \bottomrule
    \end{tabular}
\end{table*}
\begin{table*}[t]
    \caption{\small Performance comparison between total risks $\ell_{T}(\bfw_{\rm opt})$ obtained by evaluating optimal networks, with the best performing network with the minimum total risk $\min(\ell_{T}(\bfw_{\rm rand}))$ among $100$ networks that were trained on random designs $\bfw_{\rm rand}$ .}
    \label{tab:ComparingWithMins}
    \centering
    \begin{tabular}{@{}cccccc@{}}
        \toprule
        \textbf{ODE} & \textbf{Method} & \textbf{Sparsity} & $\ell_{T}(\bfw_{\rm{opt}})$ & $\min(\ell_{T}(\bfw_{\rm rand}))$& $\frac{\ell_{T}(\bfw_{\rm{opt}}) - \min(\ell_{T}(\bfw_{\rm rand})) }{\min(\ell_{T}(\bfw_{\rm rand}))} \times 100\%$ \\
        \midrule
        3-TC & Cont. w & 2 & $6.18\times10^{-2}$ &$1.35\times10^{-1}$ &$-54.38\%$ \\
        3-TC & Cont. w & 6 & $5.42\times10^{-2}$ &$6.28\times10^{-2}$ &$-13.80\%$ \\
        3-TC & Cont. w & 10 & $5.29\times10^{-2}$ &$5.32\times10^{-2}$ & $-0.62\%$ \\
        3-TC & Cont. w & 20 & $5.11\times10^{-2}$ &$5.34\times10^{-2}$ &$-4.36\%$ \\
        3-TC & TS (Binary w) & 2 &$5.78\times10^{-2}$ &$5.49\times10^{-2}$ &$-5.60\%$ \\
        3-TC & TS (Binary w) & 6 &$5.29\times10^{-2}$ &$5.49\times10^{-2}$ &$-3.65\%$ \\
        3-TC & TS (Binary w) & 10 &$5.22\times10^{-2}$ &$5.32\times10^{-2}$ &$-1.72\%$ \\
        3-TC & TS (Binary w) & 20 & $5.14\times10^{-2}$& $5.20\times10^{-2}$ & $-0.99\%$ \\
        PPM & Cont. w & 2 & $8.96\times10^{-3}$ & $9.23\times10^{-3}$& $-3.01\%$\\
        PPM & Cont. w & 4 & $6.45\times10^{-3}$ & $6.94\times10^{-3}$& $-7.02\%$\\
        PPM & Cont. w & 10 & $4.72\times10^{-3}$ &$6.01\times10^{-3}$ & $-21.53\%$ \\
        PPM & Cont. w & 20 & $4.28\times10^{-3}$ &$4.91\times10^{-3}$ & $-12.83\%$\\
        PPM & TS (Binary w) & 2 & $7.42 \times 10^{-3}$ & $1.14\times10^{-2}$ & $-35.10\%$ \\
        PPM & TS (Binary w) & 4 & $5.87 \times 10^{-3}$ & $6.14\times10^{-3}$ & $-4.47\%$ \\
        PPM & TS (Binary w) & 10 & $4.61 \times 10^{-3}$ &$5.27\times10^{-3}$  & $-12.49\%$ \\
        PPM & TS (Binary w) & 20 & $4.15 \times 10^{-3}$ &$4.33\times10^{-3}$ & $-4.31\%$ \\
        \bottomrule
    \end{tabular}
\end{table*}

\begin{table*}[t]
    \caption{\small Experimental results for parameter risks $\ell_{q}(\bfw)$. The $\ell_{q}(\bfw)$ were computed over $175k$ unseen samples for each ODE system at the given sparsity values. The corresponding Standard Error of the Mean (SEM) are shown. Also given are the mean parameter risks $\overline{\ell_{q}(\bfw_{\rm rand})}$ obtained by 100 networks trained on random designs $\bfw_{\rm rand}$. The standard deviations of $\ell_{q}(\bfw_{\rm rand})$ and the percentage difference of $\ell_{q}(\bfw_{\rm opt})$ from the random runs are similarly presented. $\bf{SP}$ refers to the sparsity of the corresponding row.}
    \label{tab:Appendix:ParameterRecoveryresults}
    \centering
    \begin{tabular}{@{}cccccccc@{}}
        \toprule
        \textbf{ODE} & \textbf{Method} & \textbf{SP} & $\ell_{q}(\bfw_{\rm{opt}})$ & \textbf{SEM}($\ell_{q}(\bfw_{\rm{opt}})$) & $\overline{\ell_{q}(\bfw_{\rm{rand}})}$ & $\frac{\ell_{q}(\bfw_{\rm{opt}}) - \overline{\ell_{q}(\bfw_{\rm{rand}})}}{\overline{\ell_{q}(\bfw_{\rm{rand}})}} \times 100\%$ & $\textbf{STD}(\ell_{q}(\bfw_{\rm{rand}})$) \\
        \midrule
        3-TC & Cont. w & 2 & $3.81\times10^{-2}$ &$9.55\times10^{-4}$ &$8.94\times10^{-2}$ &$-57.34\%$ &$2.77\times10^{-1}$ \\
        3-TC & Cont. w & 6 & $3.84\times10^{-2}$ &$1.13\times10^{-3}$ &$5.79\times10^{-2}$ &$-33.69\%$ &$1.14\times10^{-1}$ \\
        3-TC & Cont. w & 10 & $3.84\times10^{-2}$ &$1.05\times10^{-3}$ & $4.35\times10^{-2}$& $-11.87\%$& $3.69\times10^{-2}$ \\
        3-TC & Cont. w & 20 & $3.78\times10^{-2}$ &$7.57\times10^{-4}$ & $3.84\times10^{-2}$& $-1.57\%$ & $6.19\times10^{-4}$ \\
        3-TC & TS (Binary w) & 2 & $3.82\times10^{-2}$& $1.17\times10^{-3}$&$3.828\times10^{-2}$ &$-0.28\%$ &$3.94\times10^{-4}$ \\
        3-TC & TS (Binary w) & 6 & $3.80\times10^{-2}$& $1.01\times10^{-3}$&$3.819\times10^{-2}$ &$-0.43\%$ &$2.39\times10^{-4}$ \\
        3-TC & TS (Binary w) & 10 & $3.83\times10^{-2}$& $9.48\times10^{-4}$ &$3.821\times10^{-2}$ &$+0.37\%$&$2.13\times10^{-4}$\\
        3-TC & TS (Binary w) & 20 & $3.81\times10^{-2}$& $8.84\times10^{-4}$ &$3.82\times10^{-2}$ &$-0.22\%$ & $2.34\times10^{-4}$\\
        PPM & Cont. w & 2 & $1.07\times10^{-3}$ & $2.31\times10^{-5}$& $1.38\times10^{-3}$ & $-22.31\%$& $4.88\times10^{-4}$\\
        PPM & Cont. w & 4 & $5.45\times10^{-4}$ & $1.41\times10^{-5}$& $1.11\times10^{-3}$ & $-51.05\%$& $6.76\times10^{-4}$\\
        PPM & Cont. w & 10 & $4.27\times10^{-4}$ &$9.12\times10^{-6}$ & $5.54\times10^{-4}$& $-22.97\%$&$1.95\times10^{-4}$ \\
        PPM & Cont. w & 20 & $3.75\times10^{-4}$ &$1.48\times10^{-3}$ & $5.09\times10^{-4}$& $-26.44\%$& $2.10\times10^{-4}$\\
        PPM & TS (Binary w) & 2 & $7.88\times10^{-4}$ & $1.63\times10^{-5}$ & $1.60\times10^{-3}$ & $-50.76\%$ & $5.24\times10^{-4}$\\
        PPM & TS (Binary w) & 4 & $5.73\times10^{-4}$ & $1.31\times10^{-5}$ & $8.18\times10^{-4}$ & $-29.98\%$ & $3.11\times10^{-4}$\\
        PPM & TS (Binary w) & 10 & $3.58\times10^{-4}$ & $8.13\times10^{-6}$ & $6.00\times10^{-4}$ & $-40.42\%$&$3.92\times10^{-4}$ \\
        PPM & TS (Binary w) & 20 & $3.13\times10^{-4}$ & $8.13\times10^{-6}$& $3.91\times10^{-4}$ & $-19.95\%$ & $1.29\times10^{-4}$  \\
        \bottomrule
    \end{tabular}
\end{table*}

\section{Appendix: Further Experiments}
\label{appendix:c}
In this section we present results from additional experiments. In \Cref{appendix:c_A_opt} we compare the performance of a Likelihood Free Estimator (LFE) to the A-Optimal lower bound as a baseline for the case of the exponential ODE model. In \Cref{appendix: cost_comparison_conventional_approach} we compare the cost and performance of LFE to Greedy Search (\Cref{alg:greedy_search}). In \Cref{tab:lfe_training_different_gamma} we show the results of training LFEs with different data risk hyper-parameters $\gamma$ for the 3-Tissue-Compartment (3-TC) Model and the Predator-Prey Model (PPM). 

\subsection{A-Optimality and Optimal Weights}
\label{appendix:c_A_opt}
We consider a simple exponential ODE model defined by the system:
\[
\frac{dy}{dt} = \lambda * t,
\]
with the analytic solution:
\[
y = y_0 * \exp(\lambda * t).
\]

To simplify the inference problem, we take the logarithm of the analytic solution, resulting in a linearized form:
\[
\log(y) = \log(y_0) + \lambda * t.
\]
The goal of this inverse problem is to infer the parameters $\log(y_0)$ and $\lambda$ from the noisy data $d$. This leads to the following linear system:
\[
Ax = \log(d),
\]
where the matrix $A \in \mathbb{R}^{n \times m}$ with $n \geq m$ is a matrix with the first column consisting of ones and the second column containing the time points $t$ where the data $d$ are measured. In this setup, we assume that $\log(d)$ has been corrupted by additive Gaussian noise $\mathcal{N}(0, \sigma^2 I)$, where $I$ is the identity matrix. The unknown parameter vector $x = [\log(y_0), \lambda]^T$ represents the initial value $\log(y_0)$ and the decay constant $\lambda$. A lower bound on the expected variance of the parameter estimates $\mathbb{E}[MSE(x,\hat{x})]$, where $MSE$ is the mean squared error of the parameter estimates, is known analytically, and this is captured by the A-optimality criterion, and is given by:
\[
\sigma^{2} \cdot \text{trace}\left( (A^{T} W^{2} A)^{-1} \right)
\]

 We introduce binary non-negative weights $w$ associated with each time point where an entry of $1$ indicates a measurement is made at the point, and $0$ indicates that no measurement is made at the associated point. These weights are organized into a diagonal matrix $W$, transforming the system into:
\[
WAx = Wd.
\]

The lower bound for each set of candidate weights represents the variance of the system, where smaller values indicate lower uncertainty in the estimation of $x$. The expression:
\[
\mathcal{L}(w) = \sigma^{2} \cdot \text{trace}\left( (A^{T} W^{2} A)^{-1} \right) 
\]
where $\sigma^{2}$ is the additive Gaussian noise in the data $d$, and $W = \text{diag}(w)$ is the diagonal weight matrix, thus constitutes a baseline for comparison to the proposed Likelihood Free Estimator (LFE) method of obtaining the design weights $w$. 

In our experiments, the Likelihood Free Estimator (LFE) was trained by sampling the parameters $\log(y_0)$ and $\lambda$ from the following distributions:
\[
\log(y_0) \sim \mathcal{U}(0.0, 1.0), \quad \lambda \sim \mathcal{U}(-0.5, -0.01),
\]
where $\log(y_0)$ represents the logarithm of the initial value of the exponential curve, and $\lambda$ is the decay constant. The time domain $t$ of consideration was $[0,100]$, discretized to $100$ evenly spaced points. The additive Gaussian noise level $\sigma$ on $\log(d)$ was set to $0.1$. 

The Likelihood Free Estimator (LFE) network used was a feedforward neural network composed of a series of fully connected layers. The architecture is similar to Algorithm \ref{alg:net}, with the network trained exclusively on data corrupted by Gaussian noise at the constant level $\sigma=0.1$. The hidden layers had a dimensionality of 32. The SiLU (Sigmoid Linear Unit) activation function was used as the choice of non-linearity throughout the network.

We present the performance of the LFE in Table \ref{table:A_opt_performance_comparison} with respect to the appropriate A-optimality baselines for the corresponding sparsities. 

% \begin{table}[htbp]
% \centering
% \caption{The mean-squared-error (MSE $\downarrow$) of parameter recovery for A-optimality and Likelihood Free Estimator (LFE) at different sparsity levels for the exponential decay model in \ref{appendix:c_A_opt}. The additive Gaussian noise on $\log(d)$ denoted by $\sigma$, was set to $0.1$.}
% \begin{tabular}{|c|c|c|}
% \hline
% \textbf{Method} & \textbf{MSE (Sparsity = 2)} & \textbf{$\ell_{q}(\bfw_{\rm opt})$ (Sparsity = 4)} \\ \hline
% A-optimality Baseline & $0.01$ & $0.0034$ \\ \hline
% LFE  & $1.31 \times 10^{-4}$ & $4.14\times 10^{-5}$ \\ \hline
% \end{tabular}
% \label{table:A_opt_performance_comparison}
% \end{table}
\begin{table}[htbp]
\centering
\caption{The mean-squared-error (MSE $\downarrow$) of parameter recovery for A-optimality and Likelihood Free Estimator (LFE) at different sparsity levels for the exponential decay model in \ref{appendix:c_A_opt}. The additive Gaussian noise on $\log(d)$ denoted by $\sigma$, was set to $0.1$.}
\begin{tabular}{|c|c|c|}
\hline
\textbf{Method} & \textbf{MSE (Sparsity = 2)} & \textbf{$\ell_{q}(\bfw_{\rm opt})$ (Sparsity = 4)} \\ \hline
A-optimality Baseline & $1.00 \times 10^{-2}$ & $3.40 \times 10^{-3}$ \\ \hline
LFE  & $1.31 \times 10^{-4}$ & $4.14 \times 10^{-5}$ \\ \hline
\end{tabular}
\label{table:A_opt_performance_comparison}
\end{table}

\subsection{Comparison with Conventional Approaches}
\label{appendix: cost_comparison_conventional_approach}
In this sub-section, we compare parameter recovery and computational cost between the proposed Likelihood-Free Estimator (LFE) and a conventional approach, specifically Greedy Search. The Greedy Search algorithm is presented in \Cref{alg:greedy_search}. To find the optimal weights, the loss function that was minimized to obtain the parameters $q$ was $\mathcal{L}(q) = \rho\ell_{q} + \ell_{d}$ using L-BFGS as the optimization method. The regularization hyper-parameter $\rho$ was set to $0.1$ for the 3-TC model and to $1.0$ for the PPM. L-BFGS iterations were run until the loss function $\mathcal{L}(q)$ did not change within a tolerance of $10^{-6}$. For the PPM and 3-TC models, 4 L-BFGS iterations, each with 20 inner iterations were run with batch sizes of $3500$, similar to the batch sizes used in training LFEs in all our experiments. \\ 
\\
Greedy Search obtains binary design weights $\bfw$, and if continuous $\bfw$ are desired, then a further optimization run is conducted using the Adam optimizer on the chosen binary $\bfw$. For our experiments using Greedy Search, we obtain optimal design weights $\mathbf{w}$ that are continuous and compare performance with the LFE method. The cost comparison with the proposed LFE approach for training to obtain optimal $\bfw$ are presented in \Cref{tab:cost_comparison_GreedySearch}, and the performance comparisons are presented in \Cref{tab:performance_comparison_GreedySearch}. We show the inference (parameter estimation given noisy data) cost in \Cref{tab:cost_comparison_LBFGS_Inference}. 

\begin{algorithm}[h]
   \caption{Greedy Search for Optimal Design Weights}
   \label{alg:greedy_search}
\begin{algorithmic}
    \INPUT Number of discrete time points $N$, desired sparsity $sparsity$
    \STATE Initialize weight vector $\mathbf{w} \gets \mathbf{0}_{N}$
    
    \FOR{$i = 1,..., sparsity$}
       \STATE Initialize minimum loss $ \gets \infty$
       \FOR{$j = 1,..., N$}
           \IF{point $j$ is unchosen}
               \STATE Set $\mathbf{w}_j \gets 1$
               \STATE Sample true parameters $p_{true}$ from the prior distribution
               \STATE Compute parameter recovery loss: $nMSE(p_{true}, \hat{p})$ using L-BFGS
               \IF{current loss $<$ minimum loss}
                   \STATE Update minimum loss and store corresponding time point
               \ENDIF
               \STATE Set $\mathbf{w}_j \gets 0$
           \ENDIF
       \ENDFOR
       \STATE Choose the time point with the minimum loss, and permanently set $\mathbf{w}$ at the time point to 1
    \ENDFOR
    
    \STATE \textbf{Optional Continuous Optimization:}
    \IF{continuous non-binary weights are desired}
        \STATE Initialize Adam optimizer
        \STATE Initialize $\mathbf{w}_{cont}$ as the binary weights obtained above
        \FOR{$k = 1,..., {\rm num\_iters}$}
            \STATE Sample true parameters $p_{true}$ from the prior distribution
            \STATE Compute parameter recovery loss: $nMSE(p_{true}, \hat{p})$ using the continuous weights $\mathbf{w}_{cont}$
            \STATE Update $\mathbf{w}_{cont}$ using the Adam optimizer
        \ENDFOR
    \ENDIF
\end{algorithmic}
\end{algorithm}

%%%%

\begin{table}[ht]
\centering
\caption{The training cost comparison of Likelihood Free Estimator (LFE) with Greedy Search to find continuous $\bfw_{opt}$, for a batch size of $3500$. Shown are the maximum number of forward passes (F-Passes) and backward passes (B-Passes), used for automatic differentiation, and the combined wall clock time elapsed for $1$ backward and $1$ forward pass, conducted during training to infer $\bfw_{opt}$. The design weights $\bfw$ in the LFE are trained together with other neural network parameters. For Greedy Search, a further 200 iterations were conducted using PyTorch's Adam optimizer on the binary $\bfw$ obtained during the first phase, which are not included in the counts shown for F-Passes and B-Passes. Conducted on an Nvidia RTX A6000 GPU.}
\label{tab:cost_comparison_GreedySearch}
\begin{tabular}{llcccc}
\toprule
\textbf{ODE Model} & \textbf{Sparsity} & \textbf{Method} & \textbf{F-Passes (Training)} & \textbf{B-Passes (Training)} & \textbf{Time (B+F Pass) [seconds]} \\
\midrule
\multirow{4}{*}{3-TC} & \multirow{2}{*}{2} & LFE & $7000$ & $7000$ & $1.7 \times 10^{-4} \pm 1.2 \times 10^{-2}$\\
 &  & Greedy Search & $63920$ & $63920$ & $8.39 \pm 4.35$ \\
\cmidrule(lr){2-6}
 & \multirow{2}{*}{6} & LFE & $7000$ & $7000$ & $2.1 \times 10^{-4} \pm 1.3 \times 10^{-2}$\\
 &  & Greedy Search & $190800$ & $190800$ & $8.53 \pm 4.50$ \\
\midrule
\multirow{4}{*}{PPM} & \multirow{2}{*}{2} & LFE & $6200$ & $6200$&$3.74 \times 10^{-5} \pm 4.99 \times 10^{-3}$ \\
 &  & Greedy Search & $31920$ & $31920$& $1.28 \pm 1.04$\\
\cmidrule(lr){2-6}
 & \multirow{2}{*}{4} & LFE & $6200$& $6200$& $3.67 \times 10^{-5} \pm 4.99 \times 10^{-3}$ \\
 &  & Greedy Search & $63520$ & $63520$ & $1.20 \pm 1.04$ \\
\bottomrule
\end{tabular}
\end{table}

\begin{table}[ht]
\centering
\caption{Performance comparison of LFE and Greedy Search. The parameter risk $\ell_{q}(\bfw_{\rm opt})$, and data risk $\ell_{d}(\bfw_{\rm opt})$ together with their Standard Error of the Mean (SEM) are shown. The regularization hyper-parameter $\rho$ for the L-BFGS optimization runs for Greedy Search was set to $0.1$ for 3-TC and $1.0$ for PPM. The metric for the risks are the associated normalized-mean-squared-errors (nMSE ($\downarrow$)).}
\label{tab:performance_comparison_GreedySearch}
\begin{tabular}{llcccccc}
\toprule
\textbf{ODE Model} & \textbf{Sparsity} & \textbf{Method} & $\ell_{q}(\bfw_{\rm opt})$ & \textbf{SEM($\ell_{q}(\bfw_{\rm opt})$)} & $\ell_{d}(\bfw_{\rm opt})$ & \textbf{SEM($\ell_{d}(\bfw_{\rm opt})$)} \\
\midrule
\multirow{4}{*}{3-TC} & \multirow{2}{*}{2} & LFE & $3.81 \times 10^{-2}$ & $9.55 \times 10^{-4}$ & $2.37 \times 10^{-2}$ & $1.46 \times 10^{-2}$ \\
 &  & Greedy Search & $4.10 \times 10^{-2}$ & $4.05 \times 10^{-3}$ & $4.70 \times 10^{-2}$ & $3.79 \times 10^{-2}$ \\
\cmidrule(lr){2-7}
 & \multirow{2}{*}{6} & LFE & $3.84 \times 10^{-2}$ & $1.13 \times 10^{-3}$ & $1.58 \times 10^{-2}$ & $2.52 \times 10^{-4}$ \\
 &  & Greedy Search & $3.80 \times 10^{-2}$ & $9.97 \times 10^{-4}$ & $2.50 \times 10^{-2}$ & $8.04 \times 10^{-4}$ \\
\midrule
\multirow{4}{*}{PPM} & \multirow{2}{*}{2} & LFE & $1.07\times10^{-3}$ & $2.31 \times 10^{-5}$ & $7.89 \times 10^{-3}$ & $4.53 \times 10^{-4}$ \\
 &  & Greedy Search & $2.50 \times 10^{-3}$ & $5.24 \times 10^{-5}$ & $3.58 \times 10^{-2}$ & $7.11 \times 10^{-4}$ \\
\cmidrule(lr){2-7}
 & \multirow{2}{*}{4} & LFE & $5.45 \times 10^{-4}$ & $1.41 \times 10^{-5}$ & $5.91 \times 10^{-3}$ & $5.18 \times 10^{-3}$ \\
 &  & Greedy Search & $2.50 \times 10^{-3}$ & $5.27 \times 10^{-5}$ & $3.59\times 10^{-2}$ & $5.95 \times 10^{-4}$ \\
\bottomrule
\end{tabular}
\end{table}

\begin{table}[ht]
\centering
\caption{The parameter inference cost comparison of Likelihood Free Estimator (LFE) with L-BFGS to estimate parameters $q$, for a batch size of $3500$. Shown are the maximum number of forward passes (F-Passes) and backward passes (B-Passes) conducted using automatic differentiation during inference to estimate $q$. Also shown is the combined wall clock time elapsed for 1 backward and 1 forward pass (for LFE this is the time elapsed for 1 forward pass as there are no backward passes during inference). Parameter estimation is carried out for fixed optimal $\bfw$ previously obtained during training. For L-BFGS, 4 iterations with 20 inner iterations were necessary for convergence to within a tolerance of $10^{-6}$. Conducted on an Nvidia RTX A6000 GPU.}
\label{tab:cost_comparison_LBFGS_Inference}
\begin{tabular}{llcccc}
\toprule
\textbf{ODE Model} & \textbf{Sparsity} & \textbf{Method} & \textbf{F-Passes (Inference)} & \textbf{B-Passes (Inference)} & \textbf{Time (B+F Pass) [seconds]} \\
\midrule
\multirow{4}{*}{3-TC} & \multirow{2}{*}{2} & LFE & $1$ & $0$ & $1.42 \times 10^{-4} \pm 3.60 \times 10^{-4}$\\
 &  & L-BFGS & $80$ & $80$ & $8.39 \pm 4.35$ \\
\cmidrule(lr){2-6}
 & \multirow{2}{*}{6} & LFE & $1$ & $0$ & $1.93 \times 10^{-4} \pm 2.94 \times 10^{-4}$\\
 &  & L-BFGS & $80$ & $80$ & $8.53 \pm 4.50$ \\
\midrule
\multirow{4}{*}{PPM} & \multirow{2}{*}{2} & LFE & $1$ & $0$&$2.34 \times 10^{-5} \pm 2.99 \times 10^{-3}$ \\
 &  & L-BFGS & $80$ & $80$& $1.28 \pm 1.04$\\
\cmidrule(lr){2-6}
 & \multirow{2}{*}{4} & LFE & $1$& $0$& $2.40 \times 10^{-5} \pm 2.99 \times 10^{-3}$ \\
 &  & L-BFGS & $80$ & $80$ & $1.20 \pm 1.04$ \\
\bottomrule
\end{tabular}
\end{table}

\begin{table}[ht]
\centering
\caption{Results of training LFE with different data risk hyper-parameter \(\gamma\) for the 3-TC model (Sparsity=6) and PPM model (Sparsity=4). The nMSE($\downarrow$) corresponding to the parameter risk \(\ell_{p}\) and the data risk \(\ell_{d}\) are shown for each configuration.}
\label{tab:lfe_training_different_gamma}
\begin{tabular}{llcccc}
\toprule
\textbf{ODE Model} & \textbf{Sparsity} & \(\gamma\) & \(\ell_{p}\) & \(\ell_{d}\) \\
\midrule
\multirow{6}{*}{3-TC} & \multirow{6}{*}{6} & $0$ & $3.76 \times10^{-2}$ & $2.22\times10^{-2}$ \\
 &  & $1$ & $3.84\times10^{-2}$ & $1.58\times10^{-2}$ \\
 &  & $10^1$ & $3.83\times10^{-2}$ & $1.83\times10^{-2}$ \\
 &  & $10^2$ & $3.83\times10^{-2}$ & $1.79\times10^{-2}$ \\
 &  & $10^3$ & $3.85\times10^{-2}$ & $2.84\times10^{-2}$ \\
 &  & $10^4$ & $4.08\times10^{-2}$ & $1.80\times10^{-2}$ \\
\midrule
\multirow{6}{*}{PPM} & \multirow{6}{*}{4} & $0$ & $5.56\times10^{-4}$ & $6.67\times10^{-3}$ \\
 &  & $1$ & $5.45\times10^{-4}$ & $5.91\times10^{-3}$ \\
 &  & $10^1$ & $1.11 \times 10^{-3}$ & $1.68 \times 10^{-2}$ \\
 &  & $10^2$ & $1.90 \times 10^{-3}$ & $3.80 \times 10^{-2}$ \\
 &  & $10^3$ & $2.75 \times 10^{-3}$ & $6.06\times10^{-2}$ \\
 &  & $10^4$ & $1.14 \times 10^{-2}$ & $9.61 \times 10^{-2}$ \\
\bottomrule
\end{tabular}
\end{table}

\end{document}